\documentclass[review]{elsarticle}

\usepackage{lineno,hyperref}
\usepackage[svgnames]{xcolor}
\definecolor{ocre}{RGB}{10,10,10}
\usepackage{caption}
\usepackage[font={color=ocre}]{caption}
\usepackage{array}
\usepackage{amssymb}
\usepackage{amsmath}
\usepackage{subcaption}
\usepackage{adjustbox}
\usepackage{multirow}
\modulolinenumbers[5]

\journal{Journal of \LaTeX\ Templates}

\newcommand\MyBox[2]{
  \fbox{\lower0.75cm
    \vbox to 1.7cm{\vfil
      \hbox to 1.7cm{\hfil\parbox{1.4cm}{#1\\#2}\hfil}
      \vfil}%
  }%
}

\begin{document}
\begin{frontmatter}

% Main title of the paper
\title{Using Auxiliary Information for Person Re-Identification - A Tutorial Overview}  

\author[label1]{Tharindu~Fernando\corref{cor1}}
\ead{t.warnakulasuriya@qut.edu.au}

 \author[label1]{ Clinton~Fookes}
\ead{c.fookes@qut.edu.au}

\author[label1]{Sridha~Sridharan}
\ead{s.sridharan@qut.edu.au}
 
 \author[label2]{Dana~Michalski}
\ead{dana.michalski@defence.gov.au}

 \cortext[cor1]{Corresponding author at: Signal Processing, Artificial Intelligence and Vision Technologies (SAIVT), Queensland University of Technology}
 \address[label1]{Signal Processing, Artificial Intelligence and Vision Technologies (SAIVT), Queensland University of Technology.}
  \address[label2]{Defence Science and Technology Group.}

\begin{abstract}
Person re-identification (re-id) is a pivotal task within an intelligent surveillance pipeline and there exist numerous re-id frameworks that achieve satisfactory performance in challenging benchmarks. However, these systems struggle to generate acceptable results when there are significant differences between the camera views, illumination conditions, or occlusions. This result can be attributed to the deficiency that exists within many recently proposed re-id pipelines where they are predominately driven by appearance-based features and little attention is paid to other auxiliary information that could aid the re-id. In this paper, we systematically review the current State-Of-The-Art (SOTA) methods in both uni-modal and multimodal person re-id. Extending beyond a conceptual framework, we illustrate how the existing SOTA methods can be extended to support these additional auxiliary information and quantitatively evaluate the utility of such auxiliary feature information, ranging from logos printed on the objects carried by the subject or printed on the clothes worn by the subject, through to his or her behavioural trajectories. To the best of our knowledge, this is the first work that explores the fusion of multiple information to generate a more discriminant person descriptor and the principal aim of this paper is to provide a thorough theoretical analysis regarding the implementation of such a framework. In addition, using model interpretation techniques, we validate the contributions from different combinations of the auxiliary information versus the original features that the SOTA person re-id models extract. We outline the limitations of the proposed approaches and 
propose future research directions that could be pursued to advance the area of multi-modal person re-id. 
\end{abstract}
\end{frontmatter}

%\linenumbers

\section{Introduction}\label{sec:intro}
There exists a growing demand for intelligent surveillance systems and person re-identification (re-id) has become a prominent research area to support this demand. Person re-id solves the challenging task of recognising people across different camera views where there exists significant variation between viewpoints, poses, and illumination conditions. In addition, the subjects could change their attire or the objects they carry which further accentuates the challenges. Moreover, in a practical surveillance application, there may be hundreds of distinct cameras that span across a large geographic area, which will provide thousands of candidates for matching, leading to an extremely difficult task. In addition, the camera network may not have perfect coverage across the whole surveillance area resulting in blind regions. As such if the subject of interest stops for a longer period in such blind spots, retrieving candidate matches in real-time would be challenging in a practical application.

In addition to security surveillance applications, including locating a suspected criminal or locating a missing child, there are numerous commercial application of person re-id. For instance, businesses utilise person re-id frameworks to capture the customer behavioural patterns as well as customer traffic in different store areas to better allocate resources. Furthermore, person re-id can be considered the backbone of multi-target multi-camera tracking which facilitates the connection of trajectories captured from distinct camera views together.

Person re-id is an area actively researched, however, due to the breadth of challenges that it poses, it is far from being solved. An abundant number of works have focused on designing novel loss functions to promote the learning of discriminative features from the network inputs \cite{wu2018exploit, zheng2019re} while some researchers have focused on using attention mechanism to promote the localisation of crucial body parts for person re-id \cite{tay2019aanet}. Another line of work has focused on alleviating the information shortage in the 2-D frame input and has introduced novel neural network architectures such as 3-D convolutional networks \cite{liao2018video}, two stream siamese convolutional neural networks \cite{chung2017two}, and spatial-temporal attention structures \cite{fu2019sta} to extract salient information cues from video streams. Moreover, Generative Adversarial Networks (GANs) have also been readily exploited to bridge the domain gap which hinders the person re-id \cite{huang2020real, chen2019instance}. Specifically, the GANs have shown good performance in preserving identity-related attributes (self-similarity) while eliminating the dissimilarities introduced due to the domain differences. 

Other researched efforts have focused on cross-modality similarity learning, especially between the visual and textual modalities for applications where text deceptions are available \cite{liu2021spatial}. These works consider the disentanglement of the domain-specific representations (between textual and visual modalities) such that a person search can be performed with natural language queries. In addition, cross-modality learning is also utilised for capturing the information from different sensors such as thermal infrared images or depth maps. For instance, in the day-light surveillance setting, visible-light cameras provide the primary information while in dark environments thermal infrared images are the main data source. As such, several works \cite{wu2021discover, wang2020cross} have focused on learning visual representations across these information sources. However, existing works have mainly focused on methods to eliminate the domain discrepancies and align the different information sources, while investigation on the use of different auxiliary information sources that can be used to provide complementary information or methods to combine such auxiliary information has not hitherto investigated in depth \cite{ming2022deep}.

\subsection{Why auxiliary information in person re-identification?}

As mentioned earlier only a limited number of works have utilised information sources other than the visual modality for matching persons across videos.% and among those the main focus has been mitigating the cross-domain discrepancies while the learning from auxiliary information sources has been overlooked.

\begin{figure*}[htbp]
    \centering
    \includegraphics[width=\textwidth]{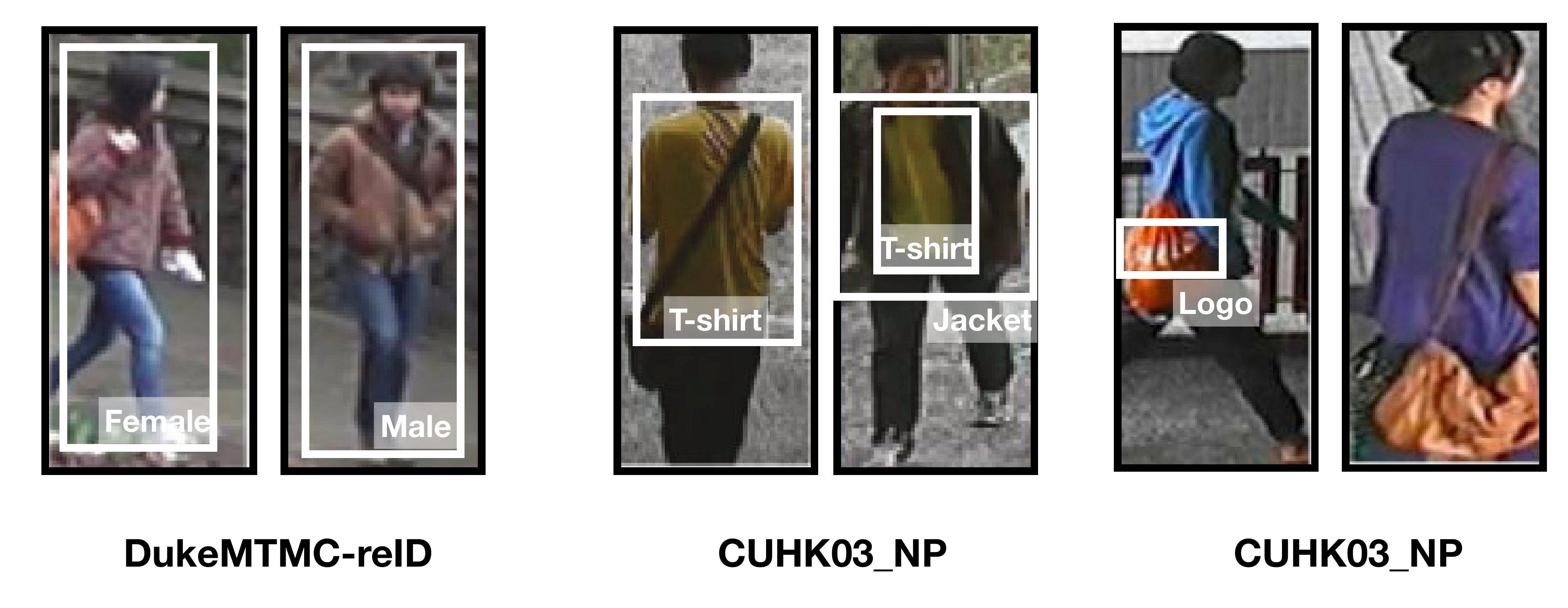}
    \caption{Motivation: Erroneous person matching that state-of-the-art methods output in public datasets (Duke Multi-Tracking Multi-Camera ReIDentification (DukeMTMC-reID) dataset \cite{ristani2016performance} and Chinese University of Hong Kong Re-identification (CUHK03-NP) dataset \cite{li2014deepreid}). The detection of auxiliary attributes such as age and gender, clothing types and logos in the objects in their possession and can potentially reduce these errors.}
    \label{fig:motivation}
\end{figure*}

Current state of the art deep learning based person re-id systems use either global features of the whole body or local features extracted from parts of the body or combination of both techniques.  However, in may applications where person re-id is deployed, the accuracy can be improved by specifically exploiting other auxiliary information that are often available such as such as tattoos in the body,  logos in the worn clothing  or in the items on possessions of an individual,  age/gender,  behaviour-based trajectories and speech,  as these  could provide additional, complementary information for identification. Our extensive search of the recent literature in this area  has revealed that very little or no research has been conducted in the use of the auxiliary information to enhance person re-id accuracies. For instance in Fig. \ref{fig:motivation} we present some of the erroneous person matching that state-of-the-art methods generate for Duke Multi-Tracking Multi-Camera ReIDentification (DukeMTMC-reID) dataset \cite{ristani2016performance} and Chinese University of Hong Kong Re-identification (CUHK03-NP) dataset \cite{li2014deepreid}. When analysing these outputs it is clearly evident that these errors can be eliminated by effectively exploiting the auxiliary attributes such as age and gender, clothing types, and logos in the objects in their possession in addition to the primary features that the re-id system learns. The aim of our paper is to address this research gap and provide a tutorial overview on the study of the extent to which these auxiliary information can be leveraged to learn a more discriminative identity representation for person re-id. 

\subsection{What auxiliary information sources can we consider?}

For this study we consider state-of-the-art person re-id systems and investigate the improvements in performance that can be obtained by exploiting specific auxiliary information that are already available in the data. For instance, estimates of age, and gender characteristics can be used to significantly reduce the search space while tattoos or logos in the objects that the subject carries (if exists) can be seen as one of the prominent information sources with the ability to provide highly discriminative attributes. Moreover, behavioral traits such as the historical trajectory of the person of interests, gait as well as voice characteristics, where available, could also be leveraged to generate identity clues. In addition, clothing type and clothing accessories could also provide informative cues for the person matching task. Even though several other auxiliary information may be available, we restrict our study to the use of age, gender, tattoos, clothing type, logos in the objects that the subject carries, behavioral traits from trajectories, and voice auxiliary information considering their prominence and abundance of availability in public benchmarks. In addition, a key objective of this study is to quantitatively evaluate the impact of the above complementary features on the person re-id task.

\subsection{Our Contributions}

In this paper, we introduce the concept of auxiliary information sources for the person re-id task and investigate the different types of information sources that can be exploited to improve performance. Going beyond a theoretical introduction we provide a tutorial-type analysis of these information sources, which outlines the extraction of these feature as well as the fusion of these auxiliary features into the existing person re-id frameworks. Moreover, in our experimental evaluations, we quantitatively and qualitatively analyse the performance of state-of-the-art person re-id systems in the presence and absence of auxiliary information sources and further interpret how the auxiliary information sources contribute to the person identification using model interpretation techniques. In addition, this paper details the limitations of existing multi-modal person re-id methods and the introduced auxiliary information sources, and proposes new research directions towards more generalisable and robust re-identification pipelines. 

\subsection{Organisation} In Sec. \ref{sec:related_works} we summarise the recent developments within the person re-id domain, which comprises state-of-the-art methods that leverage novel Convolution Neural Network (CNN) architectures, methods that devise novel attention schemes, learning methodologies, and domain adaptation techniques to achieve improved robustness. Furthermore, we summarise recent transformer-based developments within the person re-id space and the frameworks that specifically focus on the utilisation of multi-auxiliary features in the person re-id setting. Sec. \ref{sec:evaluation} provides a tutorial on formulating feature extractors to capture different auxiliary information and how to extend the existing person re-id state-of-the-art methods to accept those auxiliary features. In Sec. \ref{sec:limitations} we outline the challenges and areas for improvement, building on the results and discussion in Sec. \ref{sec:evaluation}. Sec. \ref{sec:conclusion} contains concluding remarks.

\section{Related Work}\label{sec:related_works}
In this section, we first summarise the recent developments within the person re-id domain, which comprises (i) state-of-the-art methods that leverage novel Convolution Neural Network (CNN) architectures, (ii) methods that devise novel attention schemes, (iii) learning methodologies, and (iv) domain adaptation techniques to achieve improved robustness. Furthermore, we summarise recent transformer-based developments within the person re-id space. In Sec. \ref{sec:multimodal_reid_review} we systematically review the frameworks that specifically focus on the utilisation of auxiliary information in the person re-id setting. 

\subsection{State-of-the-art in person re-identification}
The majority of the existing person re-id architectures capture the inspiration from CNN-based object classification problem \cite{zhou2019omni}. For instance, in \cite{guo2018efficient} the authors motivate the need for multi-level similarity learning and propose a novel architecture where both low-level visual features (that are captured from bottom convolutional layers) and more semantical information (which are retrieved by the higher layers) are incorporated into the similarity learning framework. Specifically, this architecture is a fully convolutional, Siamese network-based design. In the Siamese network structure, a pair of images are fed into the network and the embeddings retrieved from the network for these two images are used to determine whether they are matched. The authors demonstrate that this similarity computation is a critical component with respect to the matching accuracy and illustrate the deficiencies with the existing product of horizontal stripes \cite{li2014deepreid} and pixel to neighbors \cite{ahmed2015improved} based similarity computation methodologies. In particular, the horizontal stripes-based method is highly sensitive to the spatial alignment between the two input images, while the performance of the pixel to neighbors approach degrades when there is a significant mismatch between the sales of the two images. As such, the authors of \cite{guo2018efficient} propose to formulate the similarity between the two input images as the convolution between the extracted part from one image (filter) and the other whole image (signal). Therefore, the similarity computation can go beyond a restricted search window (i.e. neighbourhood) and can compensate for large translations and scale differences. In a similar line of work a novel CNN architecture for compensating the colour differences across cameras is proposed in \cite{wang2018person}. This network, which the authors named `BraidNet' has a cascaded convolutional structure that computes crossly sums of the features extracted from two input images. %Fig. \ref{fig:wang2018person_drawing} visually illustrate this concept. 
The authors demonstrate that the proposed structure has the ability to extract augmented features which are more robust to misalignments and colour differences. Moreover, a novel online batch generating strategy is also proposed in \cite{wang2018person}, which the authors named Sample Rate Learning (SRL). This SRL strategy has the ability to automatically increase the expected loss in each batch by dynamically adjusting the ratio of positive and negative samples in a batch.

% \begin{figure}
%     \centering
%     \includegraphics[width=\linewidth]{images/wang2018person_drawing.pdf}
%     \caption{Visual illustration of the cascaded convolutional structure in \cite{wang2018person}. 3D boxes are feature maps and Conv P/Q indicates convolution layers with P/Q size kernel.}
%     \label{fig:wang2018person_drawing}
% \end{figure}

In a different line of work, an approach specifically for video-based person re-id is proposed in \cite{subramaniam2019co}. In particular, the authors observe that when aggregating the features across the frames, features that are noisy and none irrelevant could also be aggregated into the video descriptor. There exist numerous attention-driven approaches to mitigate this issue. For instance, Li et. al \cite{li2018diversity} propose to locate distinctive body parts using spatial attention, and temporal attention is leveraged to map those across frames. In another work, \cite{wang2016person}, discriminative regions are automatically chosen using attention and simultaneously incorporated into the video descriptor. However, the authors of \cite{subramaniam2019co} argue that these methodologies are sub-optimal as they are highly dependent on the clarity of the Spatio-temporal information in the video and would fail when the person of interest is occluded. To this end, a co-segmentation-based attention network is proposed in \cite{subramaniam2019co} where objects are segmented by identifying the common objects that are available in two or more images. Therefore, this method is more resilient to background clutter and noise. 

\begin{figure*}
    \centering
    \includegraphics[width=.85\textwidth]{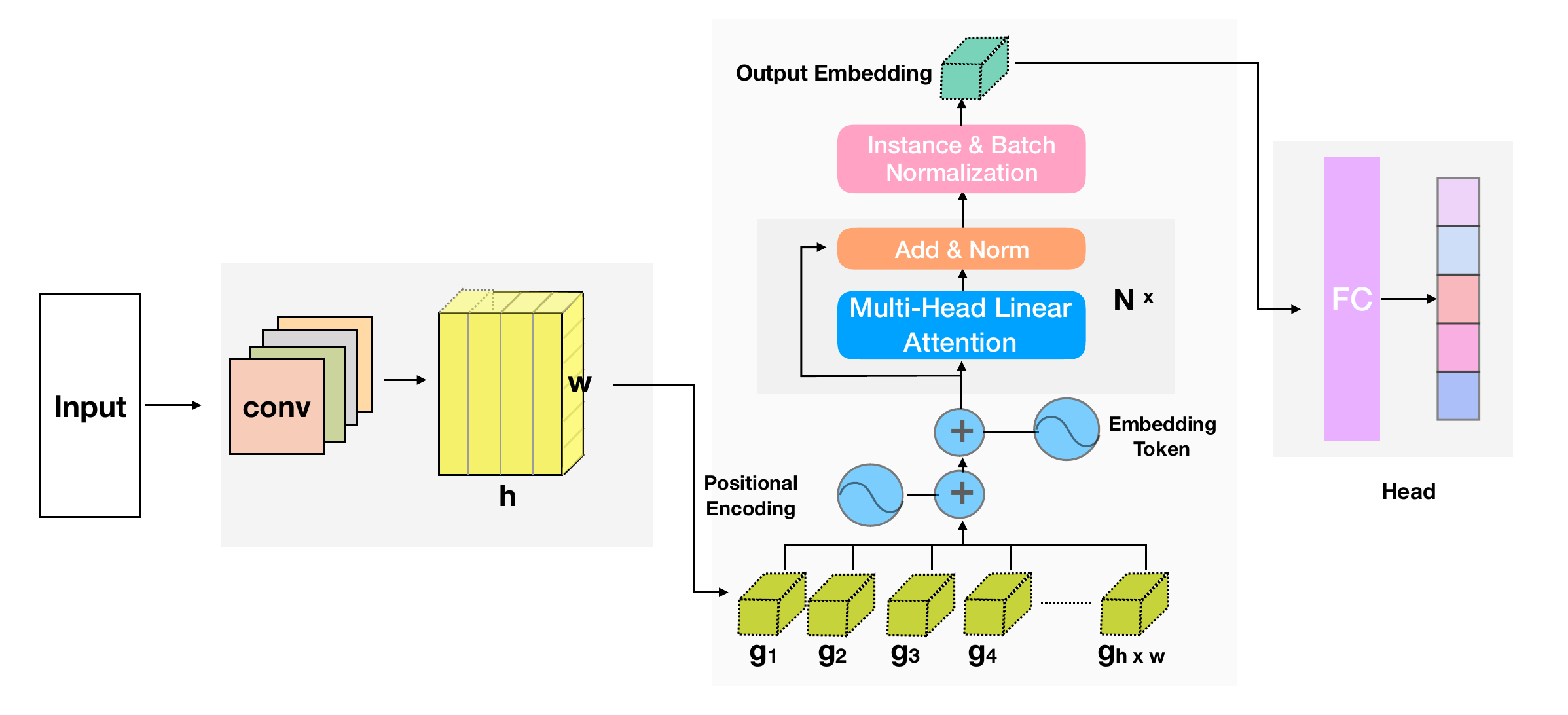}
    \caption{Overview of the transformer architecture in \cite{wu2021person} which is composed of context-based transformer encoder and part prototype-based transformer decoder. Recreated from \cite{wu2021person}.}
    \label{fig:transformer_2}
\end{figure*}

In a similar line of work a part-aware transformer network for occluded person re-id is proposed in \cite{li2021diverse}. The authors observe that in crowded places such as airports, railway stations, and hospitals, person re-id is particularly challenging due to the occlusions and design an architecture that specifically solves this problem. The authors argue that attention-based human body parts detection methods fail to capture information from the background region of the person of interest, such as his or her belongings, and focus more on locating discriminative body parts. To alleviate this issue a novel transformer-based architecture is proposed in \cite{li2021diverse}. Specifically, two transformer components are utilised, the pixel context-based transformer encoder and part prototype-based transformer decoder. The former architecture utilises self-attention to capture image context information by aggregating pixels with similar appearances. Then the part prototype-based transformer decoder utilises the aggregated pixels as the key and learns to generate body part-aware masks to extract the discriminative human body parts. The authors propose to optimise the transformer encoder-decoder jointly and learn part prototypes to the whole dataset and perform robust person re-id of occluded pedestrians. Another application of transformer networks for person re-id is available in \cite{wu2021person}. Concretely, this architecture has two main trainable components, a part feature extractor, and a transformer-based aggregator. The CNN-based part feature extractor extracts a grid-type part sequence representing the features from the individual body parts and Fig. \ref{fig:transformer_2} visually illustrates the functionality of these components.

Another notable work in the space of person re-id is available in \cite{zhou2019omni}. Specifically, the authors motivate the need for utilising a unique combination of features spanning a set of heterogeneous-scale ranging from small scales (eg. logo size) to medium scale that covers the upper body. The authors demonstrate that by combining such different scale features into the descriptor more distinctive representation of the person of interest can be generated. Concretely, the authors achieve multi-scale feature learning using a multi-stream residual function that supports different receptive field sizes. Specifically, if there are $t \in [1, \ldots T]$ scales, then the residual function can be represented as $F^t$ and they have stacked $t$ $3 \times 3$ layers and the receptive field of size can be written as $(2t +1) \times (2t + 1)$. The output of the residual is computed as the sum of the different scales up to $T$. To better fuse the representation extracted from these different scales they propose Aggregation Gates (AGs) which are trainable weights that dynamically adjust the focus given to different scales based on the input. %Fig. \ref{fig:osnet} visually illustrate the components within this bottleneck proposed in \cite{zhou2019omni}.

% \begin{figure}
%     \centering
%     \includegraphics[width=\linewidth]{images/osnet.pdf}
%     \caption{Multi-scale bottleneck architecture utilised in \cite{zhou2019omni}. AG denotes Aggregation Gate. Recreated from \cite{zhou2019omni}.}
%     \label{fig:osnet}
% \end{figure}

We would like to acknowledge the similarity between the motivation of \cite{zhou2019omni} and the proposed work. However, in this work, the authors only consider the aggregation of features across different scales. Nevertheless, they haven't considered explicit detection and extraction of discriminative features such as logos, tattoos, or clothing type. As such the proposed approach can be considered a more definite way of aggregating the multiple scales and multiple type features into the feature descriptor. Our evaluations suggest the utility of such explicit feature aggregation despite the multi-scale representations that \cite{zhou2019omni} produces.

Another approach for learning different types of features for person re-id is proposed in \cite{chang2018multi}. Specifically, the authors motivate their investigation based on the fact that discriminative factors that the CNN retrieves at different levels of abstraction should be preserved and leveraged into the feature descriptor. In particular, the proposed Multi-Level Factorisation Net (MLFN) learns multiple identity-discriminative and view-invariant visual features from the input using a set of Factor Modules (FMs) and Factor Selection Module (FSM). %Fig. \ref{fig:mlfn} depicts this concept visually. 
The FMs are sub-networks with identical architecture, however, they are tuned to learn only one factor within the feature representation. The FSMs dynamically select a set of factors (i.e. FMs) and determine what will be passed to the next level. They construct their architecture by stacking a series of blocks where each block contains several FMs and an FSM. The training process allows different FMs to specialise in different factors that are important for recognising people while FSM adjusts their saliency depending on the context. Through their evaluations, the authors demonstrate that bottom blocks in MLFN learn low-level semantics such as clothing colour while the top level learns to aggregate them into higher-level semantic representations such as objects they carry. Similar to \cite{zhou2019omni} we observe that MLFN implicitly considers the learning of an auxiliary set of visual features for person re-id. However, our evaluations suggest that the representations learned by MLFN can be further improved by explicitly modelling the auxiliary information sources. 

% \begin{figure}
%     \centering
%     \includegraphics[width=\linewidth]{images/MLFN.pdf}
%     \caption{Visual illustration of factor modules (FMs) and factor selection modules (FSM) in Multi-Level Factorisation Net \cite{chang2018multi}. Recreated from \cite{chang2018multi}.}
%     \label{fig:mlfn}
% \end{figure}

We would also like to acknowledge the breadth of works \cite{zhong2018camera, liu2019adaptive} that explore cross-domain person re-identification. The majority of these works tackle the challenging task of learning from one domain/dataset and testing the learned model across different sets of domains/datasets. It is particularly demanding considering the fact that these datasets were captured in different environments, with different camera setups and significant variances in illumination, resolution, and viewpoints. Among the different approaches that the authors of these models have undertaken, a recent trend has been the translation of the training and testing domains using GANs \cite{zhong2018camera, liu2019adaptive}. Among these works, \cite{liu2019adaptive} can be considered one of the commendable works that bridge domain gaps using a set of sub-domain transformations as opposed to the approaches that treat the domain gap as a ``black box'' and try to mitigate all the domain shifts together. The authors of \cite{liu2019adaptive} demonstrate that for different contexts, different domain discrepancy factors may hold different impacts, therefore, trying to solve all the domain shifts together in a black box setting results in sub-optimal performance. \cite{liu2019adaptive} propose a divide-and-conquer method that decomposes the cross-domain transfer problem into a set of fine-grained style transfer tasks, where each task corresponds to a certain factor that contributes to the domain shifts. These factors can include, illumination variations, resolution changes, and viewpoint variations. As such the proposed Adaptive Transfer Network (ATNet) learns an ensemble of sub-transformers that are adaptive to the context of the input image. Furthermore, a selection network is incorporated into the framework which quantifies the magnitude of the impact of these different factors in the input and allows the dynamic adaptation of ensembles. %Fig. \ref{fig:ATNet} visually illustrates how the ensemble GAN and the selection network are incorporated into the overall ATNet framework.

% \begin{figure*}
%     \centering
%     \includegraphics[width=\linewidth]{images/ATGan.pdf}
%     \caption{Visual illustration of ensemble GAN and the selection network in Adaptive Transfer Network (ATNet) \cite{liu2019adaptive}. Recreated from \cite{liu2019adaptive}.}
%     \label{fig:ATNet}
% \end{figure*}

While we acknowledge the innovations that the cross-domain person re-identification methodologies contribute, they are out of the scope of this tutorial and as such we do not utilise the cross-domain person re-id methodologies for our evaluations.

\subsection{Usage of multimodal features in person re-identification}\label{sec:multimodal_reid_review}

In addition to the unimodal person re-id frameworks that we have discussed in the previous section there exist numerous methodologies that leverage multiple information sources for person re-id. However, as discussed in Sec. \ref{sec:intro}, these works only focus on performing person identification across modalities and do not consider using cues from both information sources and building a richer feature descriptor.

Similar to unimodal person re-id within cross-modality person re-id CNNs have become the popular choice with respect to the architecture and the researchers have come up with various losses to mitigate the cross-modality discrepancies \cite{ye2018visible, ye2018hierarchical}, as well as numerous attention schemes \cite{ye2020dynamic}, disentanglement techniques \cite{choi2020hi} and modality discriminators \cite{dai2018cross} to learn a robust person descriptor which is modality invariant. Despite these attempts the misaligned features from the RGB images and the other modalities such as infrared (IR) or thermal cameras pose significant cross-modality discrepancies, generating a challenging problem to solve.

Recently, a novel framework is proposed in \cite{park2021learning} where the model encourages features from the RGB modality to reconstruct the features of the same identity in the IR space. This is achieved through the introduced CMAlign module which computes the cross-modal feature similarities. %Fig. \ref{fig:cross_modal} visually illustrates the operation of this framework. 
Specifically, it generates the probabilities of the input RGB and IR features being matched and aligns them using the proposed soft wrap operation. This operation is supported by a parameter-free person generation process which helps remove the background regions. The learning of this framework is supported by three losses, namely, the identity loss ($L_{ID}$), identity consistency loss ($L_{IC}$), and dense triplet loss ($L_{DT}$). Specifically, the $L_{ID}$ objective promotes the recognition of the identity with respect to the individual modalities while the $L_{IC}$ loss term is used to support the recognition of the same identity across different modalities, (i.e. RGB and IR images). The $L_{DT}$ facilitates the learning of discriminative person representations. This system demonstrated greater robustness due to the fact that constructed objectives force the semantically similar identities, irrespective of the modalities to be embedded nearby in the embedding space. Furthermore, the joint optimisation of dense triplet loss together with the identity losses encourages the learned features to be discriminative. 

% \begin{figure}
%     \centering
%     \includegraphics[width=\linewidth]{images/cross_modal.pdf}
%     \caption{Visual illustration of CMAlign module within the cross modality person re-id architecture in \cite{park2021learning}. Recreated from \cite{park2021learning}.}
%     \label{fig:cross_modal}
% \end{figure}

A more simplified framework for person cross-modal person re-id is proposed in \cite{kang2019person}. This framework receives either RGB or thermal images and during the learning process, it learns to reduce the discrepancies between the modalities while increasing the inter-class variation. Specifically, they have generated anchor sets using the RGB images and for the positive and negative pairs, and thermal images are pre-processed in an inter-channel and intra-channel manner. In the intra-channel case, the channels of the RGB and thermal images in the positive pairs and negative pairs are combined while in the inter-channel setting a new image is composed by arranging the RGB and thermal images of the same identity in different channels. For feature learning, a ResNet-18 CNN architecture is used. Through empirical evaluations, the authors demonstrate that this pre-processing setup contributes to the reduction of modality discrepancies.

In a different line of work, a modality translation-based approach is proposed in \cite{kniaz2018thermalgan}. Concretely, the authors propose to train a GAN to translate the appearance of the person in an RGB image into a thermal image and the generator of the GAN generates a set of probe images that anticipates how he or she would look in the thermal modality. To recognise the person in the thermal modality the authors have performed the matching between the synthesised thermal probe set and a real thermal gallery set. We point out the fact there is no existing work which considers the auxiliary information to improve person re-ID as we have  described in the introduction.

\section{Person re-id using auxiliary information fusion}\label{sec:evaluation}
\begin{figure}
    \centering
    \includegraphics[width=.95\textwidth]{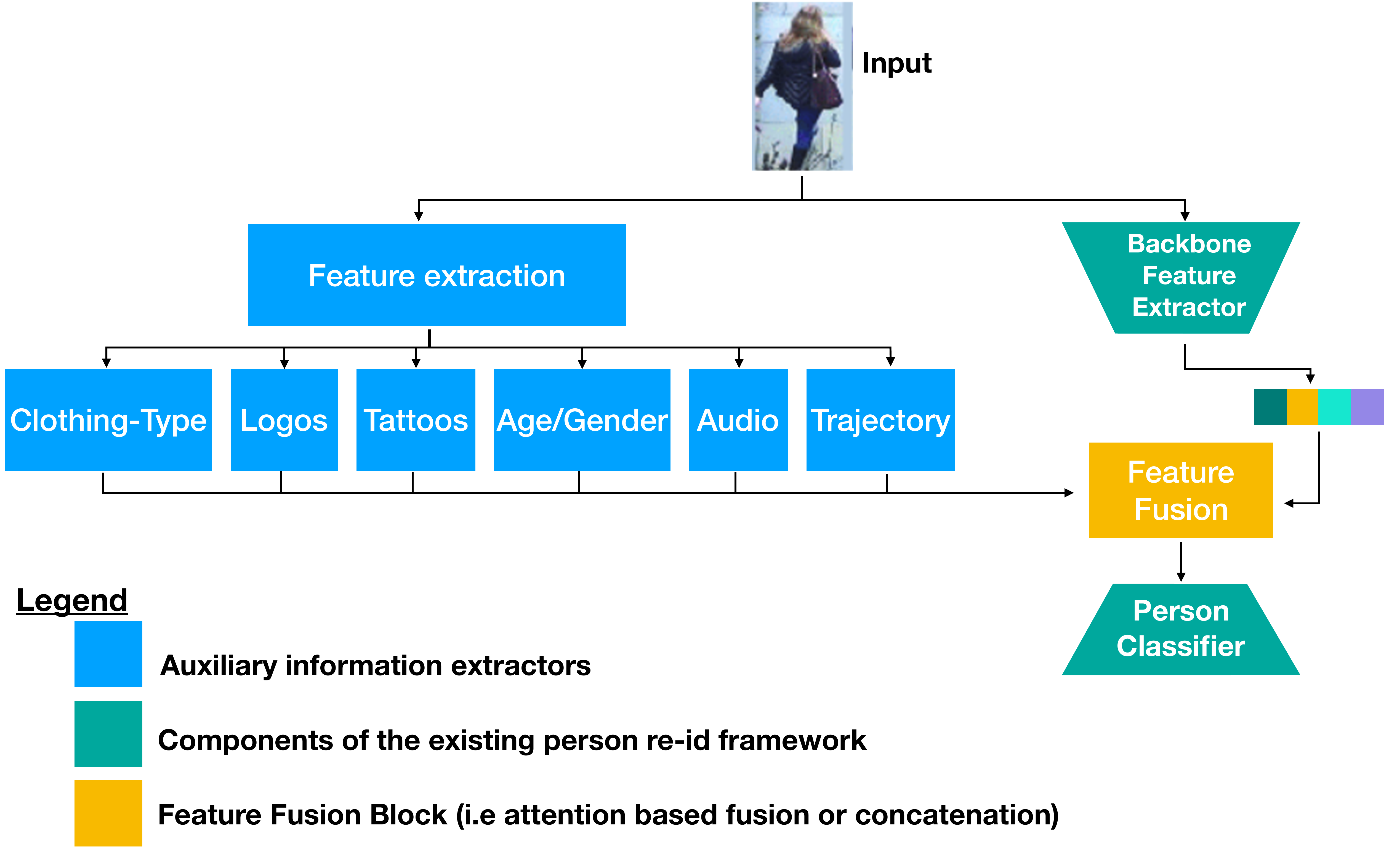}
    \caption{Overview of the proposed framework to support fusion of auxiliary information and state-of-the-art person re-id systems.}
    \label{fig:block_diagram}
\end{figure}

In this section, we first introduce the feature extractors that we design to extract the logos, age/gender, clothing types, tattoos, voice information, and behaviour trajectories. In Sec. \ref{sec:fusion} we illustrate two fusion schemes to fuse the extracted auxiliary features together with the features from the person re-id framework. Sec. \ref{sec:dataset} presents the datasets that we conducted our evaluations on and Sec. \ref{sec:results} is used to describe the quantitative and qualitative evaluation results. Overview of the framework that supports fusion of auxiliary information together with existing person re-id systems is given in Fig. \ref{fig:block_diagram}  

\subsection{Feature extractors for auxiliary information sources} 
This section presents the details of the feature extractors used and how those extracted information sources are used in the inference process. 

\subsubsection{Feature extraction using a logo detection model}
We trained a Yolo-v5 \cite{yolov5} model for logo detection using Flicker Logos-47 \cite{romberg2011scalable} dataset. In general, the Yolo object detector architecture is composed of three main components, a backbone: for extracting and aggregating the visual features at different scales, a neck: that combines the visual features that are extracted at different scales and propagate them for prediction, and a head: that accepts features from the neck and generates the bounding boxes and class predictions. Yolo is the first architecture to constitute this pipeline of predicting bounding boxes and class labels to an end-to-end framework and as such it simplifies the training procedure for custom object detection tasks. 

The Yolo-v5 which is the fifth variant in the Yolo detector family utilises several data augmentation procedures to improve its robustness, especially for small objects. Specifically, the authors propose to utilise scaling, color space adjustments, and mosaic augmentation to improve the learning process. In particular, the mosaic augmentation procedure generates synthetic additional training data by simulating random crops, and combining objects from different classes, varying the number of objects in the training images. Therefore, the output model is more robust to occlusions, translations, and unseen backgrounds. Consequently, the Yolo-v5 model has gained superior robustness making it suitable for our surveillance application. We use the public implementation provided in \url{https://github.com/ultralytics/yolov5}.

The FlickrLogos-47 dataset contains photos showing brand logos and text and is widely used for the evaluation of logo detection and recognition systems on real-world images. We used the standard training and testing splits released by the dataset authors where the training set has 833 samples while the validation set has 1,402 samples. The Yolo model was trained with a batch of 16 for 200 epochs. For training and validation curves, and the confusion matrix on the validation set please refer to Sec. 1 of the supplementary material. We test the trained model on some sample test images with logos, taken from arbitrary camera views that we downloaded from the Internet. These visualisations are provided in the supplementary material.

As our feature vector to represent the information from the logo feature we use the representation extracted from the 1st scale of the backbone. This representation provides a global overview of the input patch, as such, it can sufficiently capture the information related to the logs present in the given input. Formally, let $f_B^L$ denote the backbone of the logo detection Yolo model and $s$ denotes the scale that we have selected, then the extracted feature vector, $ \theta^L$, for logo feature, $L$, can be represented as $\theta^L = [f_B^L(X)]_s$ where $X$ is the input image patch (of the detected human in the surveillance image) and the operation $[\;]_s$ denotes the selection of the feature vector that represents scale $s$ from the feature vectors that $f_B^L$ outputs for different scales. 

% \begin{equation}
%     \theta^L = [f_B^L(X)]_s,
% \end{equation}

\subsubsection{Feature extraction using an age and gender estimation model}
We use a ResNet18 \cite{he2016deep} model pre-trained on the AFAD \cite{niu2016ordinal} dataset as our age and gender estimator We use the implementation provided in \url{https://github.com/Nebula4869/PyTorch-gender-age-estimation}. This dataset has 164,432 annotated images, where 63,680 are photos for females and 100,752 are photos for males. The participant's ages range from 15 to 40. The utilised pre-trained model achieves 93.14\% validation accuracy for the gender estimation task in the AFAD dataset. Furthermore, we tested on sample surveillance images that we obtained from WILDTRACK Seven-Camera HD Dataset \cite{chavdarova2018wildtrack} and Multiple Object Tracking (MOT) Benchmark \cite{dendorfer2021motchallenge} and this pre-trained model demonstrated satisfactory recognition performance. These sample predictions are shown in the supplementary material.

As the feature vector to represent the age and gender information we extract a 512-dimensional feature vector from the last convolution layer of the final convolutional block (i.e. layer immediately preceding the Average Pooling layer). This layer is chosen considering the fact that it comprises of most salient features for the age and gender detection tasks. We represent our ResNet encoder (without the classification layers) using $g^{A,G}$ and out input image patch as $X$, then the extracted 512-dimensional feature vector, $\theta^{A,G}$ is denoted by $\theta^{A,G} = g^{A,G}(X)$.
% \begin{equation}
%     \theta^{A,G} = g^{A,G}(X).
% \end{equation}

\subsubsection{Feature extraction using a clothing type detection model}

Similar to our logo detector a Yolo-v5 model is chosen as our clothing type detector, which is trained using the DeepFashion2 Dataset \cite{ge2019deepfashion2}. DeepFashion2 contains 491K diverse images (i.e. 391K training images and 34k validation images) of 13 popular clothing categories from both commercial shopping stores and consumers. For Yolo-v5 model training, we used randomly chosen 50k images from the training set, and for validation, we used the validation set released by the dataset authors. The Yolo model was trained with a batch of 16 for 200 epochs. Please refer to Sec. 2 of the supplementary material for training and validation curves, and the confusion matrix on the validation set. Sample outputs of the clothing type detector for the images in Fig \ref{fig:age_gender_outputs} are provided in the supplementary material.

We use the same layer as our logo detector to extract the clothing-related features. Formally, the input image patch is given by $X$ while $f_B^C$ denotes the backbone feature extractor of our Yolo-based clothing type detector model and the $s$ is the scale that we have selected from the backbone representations, then the feature vector that we extract to represent the clothing type information is given by $\theta^C$ where $\theta^C = [f_B^C(X)]_s,$ and the operation $[\;]_s$ denotes the selection of the feature vector that represents scale $s$.

% \begin{equation}
%     \theta^C = [f_B^C(X)]_s,
% \end{equation}

\subsubsection{Feature extraction using a tattoo detection model}

Considering the robustness that our Yolo detectors achieved in the previous tasks, a Yolo-v5 \cite{yolov5} model is also used for tattoo detection and the publicly available DeMSI tattoo images dataset \cite{hrkac2016tattoo} is used for training. As this dataset only has tattoo locations and tattoo classes are not annotated, we created a proxy task as the classification task of the Yolo model. If the tattoo covers more than 50px of the image then class 1 was assigned to the tattoo. Else class 0 was given.

The DeMSI tattoo images dataset has manually labelled 890 tattoo images from the ImageNet database \cite{hrkac2016tattoo} where each image contains one or more tattoos. The dataset has been annotated such that each tattoo is represented as a series of connected line segments. We created the bounding boxes for detection by defining a rectangle that covers the maximum and minimum points of this polygon. We randomly partitioned the dataset where 800 images were used to train the model and the remaining 90 images were used for validation. The Yolo model was trained with a batch of 16 for 200 epochs. Please refer to Sec. 3 of the supplementary material for training and validation curves. We tested the trained model on unseen test images from the internet to confirm its robustness and some sample outputs are given in the supplementary material.

The representations from the 1st scale of the backbone are chosen as the global representation of the tattoo information in the input patch. Hence, similar to the previous Yolo-based feature extractor we denote the backbone feature extractor of our Yolo-based tattoo detector model as $f_B^T$, $s$ the scale that we have selected, and $X$ as our input image patch, then the feature vector that we extract to represent the tattoo feature is given by $\theta^T$ and we identify this operation using $\theta^T = [f_B^T(X)]_s,$ and the operation $[\;]_s$ denotes the selection of the feature vector that represents scale $s$. Note that in all the Yolo-based feature extractors (i.e. logo, clothing type, and tattoo) we resize the input image patch to size $128 \times 128$

% \begin{equation}
%     \theta^T = [f_B^T(X)]_s,
% \end{equation}

\subsubsection{Audio feature extraction model}

We use the pre-trained SoundNet \cite{aytar2016soundnet} model trained using natural sounds as our audio feature extractor. Specifically, the SoundNet architecture has learned one-dimensional convolution filters for extracting salient acoustic representations. The authors have leveraged the natural synchronisation between vision and sound to learn sound representations across unlabelled videos. As such two visual recognition CNN networks, pre-trained using ImageNet \cite{krizhevsky2012imagenet} (object recognition) and Places \cite{zhou2014learning} (scene recognition), are employed to provide supervision using the visual features, and the audio network is trained in a student-teacher learning paradigm. Therefore, the learned acoustic features are agnostic of the final downstream task and it learns the general representation within the audio modality, making it ideal for our application.

The pre-trained SoundNet model provided by the authors of \cite{aytar2016soundnet} is trained using more than two million videos downloaded from Flickr by querying for popular tags and dictionary words. With respect to audio, Raw sound waves are used, and no pre-processing was conducted except converting the recordings to MP3s, reducing the sampling rate to 22 kHz, and converting it to single-channel audio. We use the implementation provided in \url{https://github.com/keunhong/pytorch-soundnet}. This model is capable of receiving arbitrary length audio samples in the wave format and summarising the salient content within the audio into a 2048-dimensional feature vector. This flexibility makes this model more suitable for our application. We represent the extracted 2048-dimensional feature vector as $\theta^A$, while $g^A$ is used to identify the SoundNet architecture and $M$ is the input audio recording, then the extracted feature vector for the audio modality is given by $\theta^A = g^A(M).$
% \begin{equation}
%     \theta^A = g^A(M).
% \end{equation}

\subsubsection{Feature extraction using a trajectory prediction model}

To extract the subject’s trajectories from the video data we use the Yolo-v5 \cite{yolov5} and DeepSORT \cite{wojke2017simple} algorithm. This algorithm tracks all the objects detected from the Yolo-v5 detector, as such the person class is filtered from the list of detections. The resultant trajectory is saved as a text file.

As our trajectory prediction model, we adopt a single layer LSTM \cite{hochreiter1997long} with 64 hidden units. The trajectory is regressed using a liner layer with two output units, representing the two spatial coordinates. We use this architecture instead of complex trajectory prediction methods such as \cite{fernando2018soft+, fernando2018pedestrian} due to the unavailability of neighbouring pedestrian information and sophisticated motion patterns.

The trajectory predictor is trained using the trajectories extracted from the CCTV stream of the multimodal dataset introduced in \cite{fernando2018pedestrian}. This dataset has 19,658 trajectories and 75\% of the data is randomly chosen and used to train the predictor and the remaining 25\% is used for validation. A trajectory of a length of 20-time points is chosen, and the trajectory predictor receives the first 10 points as the input, and it predicts the next 10 points. Our trajectory predictor is trained using Mean Square Error (MSE) loss with the Adam optimiser \cite{kingma2014adam} for 50 epochs. This model achieves 0.0188 an average MSE loss on the validation set.

When extracting features from the trajectory model, we extract the last hidden state of the LSTM layer which results in a 64-dimensional feature vector per trajectory. We use $g^{Trj}$ to represent our LSTM-based trajectory prediction model and $Q$ denotes the extracted trajectory of the subject, then $\theta^{Trj}$, which is the last hidden state of our trajectory predictor is obtained as $\theta^{Trj} = [g^{Trj}(Q)]'$ where the operation $[\;]'$ is used to represent the extraction of last hidden state of $g^{Trj}$. 
% \begin{equation}
%     \theta^{Trj} = [g^{Trj}(Q)]',
% \end{equation}

\subsection{Combining auxiliary information with main stream features from person identification models} \label{sec:fusion}

The next task is to combine the extracted features from our feature extractors that represent auxiliary information with the main set of features that the state-of-the-art person-reid networks utilise. As our baseline methods for evaluations we use the OSNet model introduced in \cite{zhou2019omni} and the MLFN in \cite{chang2018multi}. These methods are selected considering the similarity in the motivation that lead to the introduction of these models (i.e. incorporating different attributes of the subject into the feature descriptor) and the motivation of the proposed work, their robustness in public datasets, as well as the public availability of the implementations. 

In addition to these state-of-the-art methodologies in person-reid, we also adapt the ResNet50 \cite{he2016deep} architecture, which was originally introduced for the image recognition task, for person-reid. In addition, MobileNetV2 \cite{sandler2018mobilenetv2} was also evaluated to observe how well this lightweight counterpart of ResNet50 scales for person-reid with the help of auxiliary features. As such our evaluation include a popular image-based model and a lightweight model that can be utilised for the person-reid task as well as models specifically introduced for person-reid. We believe such comparisons across a diverse set of models will demonstrate the generalisability of our approach.

When considering the fusion of the extracted auxiliary feature information together with the main features that the person-reid framework captures, the simplest way to combine them is to concatenate them together. Specifically, let $\Theta^{reid}$ denote the primary features that the person-reid model has identified from the input  and let $\Theta^{aux}$ denotes our auxiliary features where $\Theta^{aux} = [\theta^L, \theta^{A,G}, \theta^C, \theta^T, \theta^A, \theta^{Trj}]$ and $[,]$ denotes concatenation, then we can formulated our augmented feature vector, $\Theta^*$, for the classifier of our person-reid system as $ \Theta^* =  [\Theta^{reid}, \Theta^{aux}]$.

% \begin{equation}
% \begin{split}
%     \Theta^* & =  [\Theta^{reid}, \Theta^{aux}], \\
%           & =[\Theta^{reid}, \theta^L, \theta^{A,G}, \theta^C, \theta^T, \theta^A, \theta^{Trj}].
% \end{split}
% \end{equation}
Note that for illustrative purposes in the definition of $\Theta^{aux}$ we have included all the auxiliary information that we consider, however, depending on the evaluations an arbitrary number of auxiliary information sources can be combined.

As we have selected a diverse set of information sources, with respect to the type of characteristics that it captures from the input as well as the scale that information sources is represented in the input, the information that our augmented feature vector, $\Theta^*$, captures is highly distinct. Therefore, it is naive to simply concatenate such a diverse set of information sources together as informative content can be suppressed or missed by the classifier. Moreover, all the information sources are not equally relevant in all the contexts and the model should learn to pay varying levels of attention to them based on the application. As such we experiment with a second feature fusion strategy where we pass our augmented feature vector, $\Theta^*$, through an attention scheme such that the model can learn to emphasise the salient information sources in the combined vector. Formally, let $a$ denote the attention layer (i.e a linear layer that generates attention scores) where $\alpha = a(\Theta^*)$
% \begin{equation}
%     \alpha = a(\Theta^*),
% \label{eq:attention}
% \end{equation}
and we normalise the scores using,
\begin{equation}
    \beta_{j} = \frac{\mathrm{exp}(\alpha_j)}{\sum_{r=1}^{d}\mathrm{exp}(\alpha_r)}, \quad \forall j \in d,
\end{equation}
where $d$ denotes the dimensionality of $\Theta^*$. 

Then we obtain the attention based enhanced final feature representation, $\tilde{\Theta}$, as $\tilde{\Theta} = \Theta^* \otimes \beta$
% \begin{equation}
%     \tilde{\Theta} = \Theta^* \otimes \beta,
% \end{equation}
where $\otimes$ denotes element-wise multiplication. 

Our implementations are based on the deep-person-reid python library \footnote{Publicly available from \url{https://github.com/KaiyangZhou/deep-person-reid} } that the authors of the OSNet model \cite{zhou2019omni} have released and in our evaluations, we extend this library to support the auxiliary feature extraction and the incorporation of those extracted features into the person-reid models. Depending on the utilised datasets there can be a variety of auxiliary information combinations that the reid models can receive, and as such our implementations can support an arbitrary number of features and supports extraction of the auxiliary features from both images and videos. Therefore, our extensions can be seamlessly incorporated into both image and video person re-identification datasets. 

All the components of the person-reid models are trained together with the attention layers while the auxiliary feature extractors were pre-trained. For training, all the person-reid models we use the softmax loss function due to its simplicity, and all the models were trained using Adam \cite{kingma2014adam} optimiser with a learning rate of 0.0003 for 100 epochs. For video datasets, a batch size of 5 is used while for image datasets that is set to 32.

\subsection{Datasets}\label{sec:dataset}

In this section, we discuss the details of the datasets that we have utilised in our evaluations. We have considered both image-based person-reid datasets and video-based person-reid datasets to demonstrate the viability of our approach across numerous settings as well as information streams. 

\subsubsection{Person RE-ID 2011 (PRID2011) dataset} % video based
Person RE-ID 2011 \cite{hirzer2011person} has been captured from two different static surveillance cameras, monitoring crosswalks and sidewalks. In contrast to image-based person-reid datasets, this dataset is extracted from multiple person trajectories extracted from surveillance videos, therefore, each subject that is available for identification has several different poses per person in each camera view. Furthermore, there exist between 100-150 images per subject in each camera view. Camera view A shows 475 persons, and camera view B shows 753 persons. There are 245 subjects who appear in both camera views. This dataset offers a challenging evaluating setting with significant differences in illumination conditions, backgrounds, and camera parameters. In our evaluations we utilised the dataset splits provided by the dataset authors.

\subsubsection{Duke Multi-Tracking Multi-Camera ReIDentification (DukeMTMC-reID) dataset}
The Duke Multi-Tracking Multi-Camera ReIDentification \cite{ristani2016performance} is one of the largest and most popular datasets for the evaluation of person re-id systems. This data has been collected from 8 synchronised static surveillance cameras, on Duke University's campus during periods between lectures when the pedestrian traffic is heavy. The pedestrian trajectories are manually annotated, as such the quality of the extracted subject patches is very high. The dataset comprises 16,522 training images (702 identities), 2,228 query images of the other (702 identities), and 17,661 gallery images. In addition to the large size of the dataset with respect to the number of different identities, it also offers a significant variability of crowd densities which varies from 0 to 54 people per frame. High crowd density provides an additional challenge for the person re-id as the extracted pedestrian bounding boxes can include occlusion. 

Furthermore, in contrast to many publicly available benchmarks, this dataset is not a scripted dataset which makes the subjects have various poses and carry numerous objects, including bags, backpacks, umbrellas, and bicycles. In addition, the wide field of view makes the detection of fine-grained details of the detected pedestrian particularly challenging. For our evaluations we used the dataset splits utilised by the authors of \cite{zhou2019omni}.

\subsubsection{Motion Analysis and Re-identification Set (MARS) dataset} % video based
Motion Analysis and Re-identification Set (MARS) \cite{zheng2016mars} is one of the largest video-based person reidentification datasets and it has been captured from six near-synchronized cameras placed at the Tsinghua university. Five of those cameras are HD cameras capturing data in $1080 \time 1920$ resolution, while the other camera is a standard resolution camera and captured the data in $640 \time 480$ resolution. Person trajectories are extracted by detecting the pedestrians automatically using the DPM detector \cite{felzenszwalb2010object} and tracking them using the GMMCP tracker \cite{dehghan2015gmmcp}. As such, compared to other public benchmarks the MARS dataset offers a unique evaluation setting with the existence of detection/tracking errors that are realistic in a practical application setting. 

MARS dataset consists of 1,261 different pedestrians, who are captured by at least 2 cameras. This dataset has challenging, illumination, pose and colour variations as well as 3,248 distractor pedestrians which makes it ideal for real-world evaluations. Moreover, it is 30 times larger in the number of identities compared to the PRID-2011 dataset. The authors of the dataset provide a standard evaluation protocol where subjects are evenly divided into train and test sets. We utilise the same evaluation protocol which resulted in 631 identities for training and 630 identities for testing.

\subsubsection{MSU-AVIS dataset}

Despite its tremendous viability for identification, audio is not a common modality in most of the public person reidentification benchmarks. This is mainly because most public datasets are captured in outdoor surveillance settings where it is practically difficult to capture the audio. We identify the MSU-AVIS dataset \cite{chowdhury2018msu}, which is an indoor surveillance dataset as a possible candidate for our evaluation of audio and trajectory information. This dataset is acquired using a consumer-grade camera with a built-in microphone and the video is captured at $1920 \times 1080$ resolution while the audio has a 48kHz sampling rate. This dataset has 50 subjects and has face pose, expression, and distance variations, and both clean and degraded audio. Moreover, speaker recognition is text-independent. Furthermore, the high-resolution video allows us to evaluate clothing type information simultaneously with both audio and trajectory information. Note that among the selected public datasets MSU-AVIS dataset is the only dataset that provides a complete view of the scene (instead of the cropped person view). Therefore, we were able to track the subjects across the video and extract their trajectories, which will be the inputs to the trajectory feature extractor. %Some example detections of clothing type and age/gender estimation models are provided in Fig. \ref{fig:msu_avis}.

% \begin{figure*}[htbp]
%      \centering
%      \begin{subfigure}[b]{0.45\linewidth}
%          \centering
%          \includegraphics[width=\textwidth]{images/msu_avis_1.png}
%      \end{subfigure}
%      \begin{subfigure}[b]{0.45\linewidth}
%          \centering
%          \includegraphics[width=\textwidth]{images/msu_avis_2.png}
%      \end{subfigure}
%      \caption{Testing the trained clothing type and age/gender estimation models on sample images from the MSU-AVIS dataset. }
%           \label{fig:msu_avis}
% \end{figure*} 

Due to the unavailability of the standard training and testing splits we randomly selected 25 subjects for training and the remaining for testing. When conducting our evaluations, all the baseline models and their counterparts that receives auxiliary information are trained and tested on the same training/testing splits, which ensures direct comparability. However, to ensure the robustness of the evaluations irrespective of the splits, we ran all the evaluations 5 times, randomly selecting the candidates for the splits. The average evaluation scores across the 5 runs are reported. 

\subsubsection{NBA Players dataset} % image based

All the datasets that are considered thus far are video-based datasets. However, the poor resolution of existing public datasets limits the utilisation of tattoo and logo information in our evaluations. As such, to evaluate the impact of the tattoo and logo information we have collected a new image-based person re-id dataset.

A recent survey has revealed that more than 70\% of NBA players have at least one tattoo on their body. Furthermore, these tattoos are highly unique with respect to colors and shapes which range from arm sleeves to cartoon characters and from motivational phrases to portraits. This variety makes NBA players an ideal test bed to evaluate the contribution of tattoo information when recognising a person.

With this motivation, we collected a dataset that consists of 162 popular NBA players. Our database includes players with various levels of tattoos covering their bodies. For instance, it contains players with a considerably low number of tattoos such as Cory Joseph to players like Monta Ellis who has significantly more tattoos. We utilised Google image search to construct our dataset where we queried the images of the 162 players and for each player, the top 15 images from the result were downloaded. Therefore, the image downloading process was fully automated and we didn't manually examine the collected data to ensure their quality and accuracy. The downloaded images were resized to $300 \times 300$ pixel size and we did not apply any other post-processing to the collected images. \footnote{Dataset Link to be released soon}

Due to this arbitrary acquisition of the data, there exists significant viewpoint, clothing, camera angle, and illumination variations within this dataset. Which makes it a significantly challenging dataset to achieve high re-id rates.

We randomly choose 129 player identities for training and the remaining 33 identities for testing. Similar to the MSU-AVIS dataset, the evaluations were repeated 5 times to ensure their repeatability and we report the average values across the 5 runs. 

\subsection{Results}\label{sec:results}
In this section, we provide the evaluations of the ResNet50, MobileNet, OsNet, and MLFN models and their concatenation (denoted as [Model-Name]-Concat) and attention-based (denoted as [Model-Name]-Att) feature fusion variants using PRID2011, DukeMTMC-reID, MARS, and NBA-Players datasets.

\subsubsection{Results for the PRID2011 dataset}

The evaluations of the ResNet50, MobileNet, OsNet, and MLFN models and their concatenation and attention-based feature fusion variants on the PRID2011 dataset are presented in Tab. \ref{tab:prid}. A clear utility of the additional information sources is observed. For instance, even with the addition of features from clothing type approximately a 2\% increase in the mAP metric is observed for the OsNet model while a substantial 49.9\% increase is observed for ResNet50. We observe a contribution from both clothing type and Age/Gender information. When considering the impact of the attention-based feature fusion mechanism we observe inconsistent results. For MobileNet and MLFN models, we observe a clear advantage, while in OsNet and ResNet50 the concatenation-based fusion strategy has gained superior results. We believe this is due to the structure of the network and added trainable attention parameters could lead to a significant training overhead and the limited training data availability could impact the convergence of the model. 

\begin{table*}[htbp]
\resizebox{\textwidth}{!}{
\begin{tabular}{cccccccccccccc}
\hline
\multicolumn{1}{|c|}{}                                               & \multicolumn{1}{c|}{Metric}  & \multicolumn{1}{c|}{ResNet50} & \multicolumn{1}{c|}{ResNet50-Concat} & \multicolumn{1}{c|}{ResNet50-Att} & \multicolumn{1}{c|}{MobileNet} & \multicolumn{1}{c|}{MobileNet-Concat} & \multicolumn{1}{c|}{MobileNet-Att} & \multicolumn{1}{c|}{OsNet}   & \multicolumn{1}{c|}{OsNet-Concat} & \multicolumn{1}{c|}{OsNet-Att} & \multicolumn{1}{c|}{MLFN}   & \multicolumn{1}{c|}{MLFN-Concat} & \multicolumn{1}{c|}{MLFN-Att} \\ \hline
\multicolumn{1}{|c|}{\multirow{5}{*}{PRID2011}}                      & \multicolumn{1}{c|}{mAP}     & \multicolumn{1}{c|}{29.0\%}   & \multicolumn{1}{c|}{-}               & \multicolumn{1}{c|}{-}            & \multicolumn{1}{c|}{51.0\%}    & \multicolumn{1}{c|}{-}                & \multicolumn{1}{c|}{-}             & \multicolumn{1}{c|}{89.2\%}  & \multicolumn{1}{c|}{-}            & \multicolumn{1}{c|}{-}         & \multicolumn{1}{c|}{33.2\%} & \multicolumn{1}{c|}{-}           & \multicolumn{1}{c|}{-}        \\ \cline{2-14} 
\multicolumn{1}{|c|}{}                                               & \multicolumn{1}{c|}{Rank-1}  & \multicolumn{1}{c|}{13.5\%}   & \multicolumn{1}{c|}{-}               & \multicolumn{1}{c|}{-}            & \multicolumn{1}{c|}{38.2\%}    & \multicolumn{1}{c|}{-}                & \multicolumn{1}{c|}{-}             & \multicolumn{1}{c|}{84.3\%}  & \multicolumn{1}{c|}{-}            & \multicolumn{1}{c|}{-}         & \multicolumn{1}{c|}{21.3\%} & \multicolumn{1}{c|}{-}           & \multicolumn{1}{c|}{-}        \\ \cline{2-14} 
\multicolumn{1}{|c|}{}                                               & \multicolumn{1}{c|}{Rank-5}  & \multicolumn{1}{c|}{44.9\%}   & \multicolumn{1}{c|}{-}               & \multicolumn{1}{c|}{-}            & \multicolumn{1}{c|}{64.0\%}    & \multicolumn{1}{c|}{-}                & \multicolumn{1}{c|}{-}             & \multicolumn{1}{c|}{94.4\%}  & \multicolumn{1}{c|}{-}            & \multicolumn{1}{c|}{-}         & \multicolumn{1}{c|}{44.9\%} & \multicolumn{1}{c|}{-}           & \multicolumn{1}{c|}{-}        \\ \cline{2-14} 
\multicolumn{1}{|c|}{}                                               & \multicolumn{1}{c|}{Rank-10} & \multicolumn{1}{c|}{57.3\%}   & \multicolumn{1}{c|}{-}               & \multicolumn{1}{c|}{-}            & \multicolumn{1}{c|}{76.4\%}    & \multicolumn{1}{c|}{-}                & \multicolumn{1}{c|}{-}             & \multicolumn{1}{c|}{98.9\%}  & \multicolumn{1}{c|}{-}            & \multicolumn{1}{c|}{-}         & \multicolumn{1}{c|}{58.4\%} & \multicolumn{1}{c|}{-}           & \multicolumn{1}{c|}{-}        \\ \cline{2-14} 
\multicolumn{1}{|c|}{}                                               & \multicolumn{1}{c|}{Rank-20} & \multicolumn{1}{c|}{84.3\%}   & \multicolumn{1}{c|}{-}               & \multicolumn{1}{c|}{-}            & \multicolumn{1}{c|}{89.9\%}    & \multicolumn{1}{c|}{-}                & \multicolumn{1}{c|}{-}             & \multicolumn{1}{c|}{100.0\%} & \multicolumn{1}{c|}{-}            & \multicolumn{1}{c|}{-}         & \multicolumn{1}{c|}{68.5\%} & \multicolumn{1}{c|}{-}           & \multicolumn{1}{c|}{-}        \\ \hline
\multicolumn{1}{|c|}{\multirow{5}{*}{PRID2011 + Clothing}}              & \multicolumn{1}{c|}{mAP}     & \multicolumn{1}{c|}{-}        & \multicolumn{1}{c|}{78.9\%}          & \multicolumn{1}{c|}{53.1\%}       & \multicolumn{1}{c|}{-}         & \multicolumn{1}{c|}{57.4\%}           & \multicolumn{1}{c|}{61.9\%}        & \multicolumn{1}{c|}{-}       & \multicolumn{1}{c|}{91.5\%}       & \multicolumn{1}{c|}{88.7\%}    & \multicolumn{1}{c|}{-}      & \multicolumn{1}{c|}{36.9\%}      & \multicolumn{1}{c|}{45.1\%}   \\ \cline{2-14} 
\multicolumn{1}{|c|}{}                                               & \multicolumn{1}{c|}{Rank-1}  & \multicolumn{1}{c|}{-}        & \multicolumn{1}{c|}{69.7\%}          & \multicolumn{1}{c|}{39.3\%}       & \multicolumn{1}{c|}{-}         & \multicolumn{1}{c|}{44.9\%}           & \multicolumn{1}{c|}{50.6\%}        & \multicolumn{1}{c|}{-}       & \multicolumn{1}{c|}{88.8\%}       & \multicolumn{1}{c|}{84.3\%}    & \multicolumn{1}{c|}{-}      & \multicolumn{1}{c|}{22.5\%}      & \multicolumn{1}{c|}{32.6\%}   \\ \cline{2-14} 
\multicolumn{1}{|c|}{}                                               & \multicolumn{1}{c|}{Rank-5}  & \multicolumn{1}{c|}{-}        & \multicolumn{1}{c|}{91.0\%}          & \multicolumn{1}{c|}{70.8\%}       & \multicolumn{1}{c|}{-}         & \multicolumn{1}{c|}{71.9\%}           & \multicolumn{1}{c|}{79.8\%}        & \multicolumn{1}{c|}{-}       & \multicolumn{1}{c|}{96.6\%}       & \multicolumn{1}{c|}{92.1\%}    & \multicolumn{1}{c|}{-}      & \multicolumn{1}{c|}{58.4\%}      & \multicolumn{1}{c|}{56.2\%}   \\ \cline{2-14} 
\multicolumn{1}{|c|}{}                                               & \multicolumn{1}{c|}{Rank-10} & \multicolumn{1}{c|}{-}        & \multicolumn{1}{c|}{95.5\%}          & \multicolumn{1}{c|}{79.8\%}       & \multicolumn{1}{c|}{-}         & \multicolumn{1}{c|}{82.0\%}           & \multicolumn{1}{c|}{87.6\%}        & \multicolumn{1}{c|}{-}       & \multicolumn{1}{c|}{97.8\%}       & \multicolumn{1}{c|}{96.6\%}    & \multicolumn{1}{c|}{-}      & \multicolumn{1}{c|}{68.5\%}      & \multicolumn{1}{c|}{70.8\%}   \\ \cline{2-14} 
\multicolumn{1}{|c|}{}                                               & \multicolumn{1}{c|}{Rank-20} & \multicolumn{1}{c|}{-}        & \multicolumn{1}{c|}{100.0\%}         & \multicolumn{1}{c|}{88.8\%}       & \multicolumn{1}{c|}{-}         & \multicolumn{1}{c|}{92.1\%}           & \multicolumn{1}{c|}{97.8\%}        & \multicolumn{1}{c|}{-}       & \multicolumn{1}{c|}{98.9\%}       & \multicolumn{1}{c|}{100.0\%}   & \multicolumn{1}{c|}{-}      & \multicolumn{1}{c|}{75.3\%}      & \multicolumn{1}{c|}{82.0\%}   \\ \hline
\multicolumn{1}{|c|}{\multirow{5}{*}{PRID2011 + Clothing + Age/Gender}} & \multicolumn{1}{c|}{mAP}     & \multicolumn{1}{c|}{-}        & \multicolumn{1}{c|}{78.3\%}          & \multicolumn{1}{c|}{57.2\%}       & \multicolumn{1}{c|}{-}         & \multicolumn{1}{c|}{62.5\%}           & \multicolumn{1}{c|}{64.5\%}        & \multicolumn{1}{c|}{-}       & \multicolumn{1}{c|}{93.8\%}       & \multicolumn{1}{c|}{91.9\%}    & \multicolumn{1}{c|}{-}      & \multicolumn{1}{c|}{52.4\%}      & \multicolumn{1}{c|}{60.4\%}   \\ \cline{2-14} 
\multicolumn{1}{|c|}{}                                               & \multicolumn{1}{c|}{Rank-1}  & \multicolumn{1}{c|}{-}        & \multicolumn{1}{c|}{66.3\%}          & \multicolumn{1}{c|}{41.6\%}       & \multicolumn{1}{c|}{-}         & \multicolumn{1}{c|}{49.4\%}           & \multicolumn{1}{c|}{50.6\%}        & \multicolumn{1}{c|}{-}       & \multicolumn{1}{c|}{92.1\%}       & \multicolumn{1}{c|}{88.8\%}    & \multicolumn{1}{c|}{-}      & \multicolumn{1}{c|}{38.2\%}      & \multicolumn{1}{c|}{48.3\%}   \\ \cline{2-14} 
\multicolumn{1}{|c|}{}                                               & \multicolumn{1}{c|}{Rank-5}  & \multicolumn{1}{c|}{-}        & \multicolumn{1}{c|}{89.9\%}          & \multicolumn{1}{c|}{76.4\%}       & \multicolumn{1}{c|}{-}         & \multicolumn{1}{c|}{75.3\%}           & \multicolumn{1}{c|}{82.0\%}        & \multicolumn{1}{c|}{-}       & \multicolumn{1}{c|}{96.6\%}       & \multicolumn{1}{c|}{95.5\%}    & \multicolumn{1}{c|}{-}      & \multicolumn{1}{c|}{71.9\%}      & \multicolumn{1}{c|}{75.3\%}   \\ \cline{2-14} 
\multicolumn{1}{|c|}{}                                               & \multicolumn{1}{c|}{Rank-10} & \multicolumn{1}{c|}{-}        & \multicolumn{1}{c|}{97.8\%}          & \multicolumn{1}{c|}{86.5\%}       & \multicolumn{1}{c|}{-}         & \multicolumn{1}{c|}{88.8\%}           & \multicolumn{1}{c|}{91.0\%}        & \multicolumn{1}{c|}{-}       & \multicolumn{1}{c|}{96.6\%}       & \multicolumn{1}{c|}{96.6\%}    & \multicolumn{1}{c|}{-}      & \multicolumn{1}{c|}{80.9\%}      & \multicolumn{1}{c|}{87.6\%}   \\ \cline{2-14} 
\multicolumn{1}{|c|}{}                                               & \multicolumn{1}{c|}{Rank-20} & \multicolumn{1}{c|}{-}        & \multicolumn{1}{c|}{100.0\%}         & \multicolumn{1}{c|}{89.9\%}       & \multicolumn{1}{c|}{-}         & \multicolumn{1}{c|}{95.5\%}           & \multicolumn{1}{c|}{96.6\%}        & \multicolumn{1}{c|}{-}       & \multicolumn{1}{c|}{97.8\%}       & \multicolumn{1}{c|}{100.0\%}   & \multicolumn{1}{c|}{-}      & \multicolumn{1}{c|}{89.9\%}      & \multicolumn{1}{c|}{95.5\%}   \\ \hline
\multicolumn{1}{l}{}                                                 & \multicolumn{1}{l}{}         & \multicolumn{1}{l}{}          & \multicolumn{1}{l}{}                 & \multicolumn{1}{l}{}              & \multicolumn{1}{l}{}           & \multicolumn{1}{l}{}                  & \multicolumn{1}{l}{}               & \multicolumn{1}{l}{}         & \multicolumn{1}{l}{}              & \multicolumn{1}{l}{}           & \multicolumn{1}{l}{}        & \multicolumn{1}{l}{}             & \multicolumn{1}{l}{}          \\
\multicolumn{1}{l}{}                                                 & \multicolumn{1}{l}{}         & \multicolumn{1}{l}{}          & \multicolumn{1}{l}{}                 & \multicolumn{1}{l}{}              & \multicolumn{1}{l}{}           & \multicolumn{1}{l}{}                  & \multicolumn{1}{l}{}               & \multicolumn{1}{l}{}         & \multicolumn{1}{l}{}              & \multicolumn{1}{l}{}           & \multicolumn{1}{l}{}        & \multicolumn{1}{l}{}             & \multicolumn{1}{l}{}          \\
\multicolumn{1}{l}{}                                                 & \multicolumn{1}{l}{}         & \multicolumn{1}{l}{}          & \multicolumn{1}{l}{}                 & \multicolumn{1}{l}{}              & \multicolumn{1}{l}{}           & \multicolumn{1}{l}{}                  & \multicolumn{1}{l}{}               & \multicolumn{1}{l}{}         & \multicolumn{1}{l}{}              & \multicolumn{1}{l}{}           & \multicolumn{1}{l}{}        & \multicolumn{1}{l}{}             & \multicolumn{1}{l}{}          \\
\multicolumn{1}{l}{}                                                 & \multicolumn{1}{l}{}         & \multicolumn{1}{l}{}          & \multicolumn{1}{l}{}                 & \multicolumn{1}{l}{}              & \multicolumn{1}{l}{}           & \multicolumn{1}{l}{}                  & \multicolumn{1}{l}{}               & \multicolumn{1}{l}{}         & \multicolumn{1}{l}{}              & \multicolumn{1}{l}{}           & \multicolumn{1}{l}{}        & \multicolumn{1}{l}{}             & \multicolumn{1}{l}{}          \\
\multicolumn{1}{l}{}                                                 & \multicolumn{1}{l}{}         & \multicolumn{1}{l}{}          & \multicolumn{1}{l}{}                 & \multicolumn{1}{l}{}              & \multicolumn{1}{l}{}           & \multicolumn{1}{l}{}                  & \multicolumn{1}{l}{}               & \multicolumn{1}{l}{}         & \multicolumn{1}{l}{}              & \multicolumn{1}{l}{}           & \multicolumn{1}{l}{}        & \multicolumn{1}{l}{}             & \multicolumn{1}{l}{}         
\end{tabular}}
\vspace{-15mm}
\caption{Evaluation results of the Person RE-ID 2011 (PRID2011) \cite{hirzer2011person} dataset.}
\label{tab:prid}
\end{table*}

\subsubsection{Results for the DukeMTMC-reID dataset}

The evaluation results of the DukeMTMC-reID dataset are presented in Tab. \ref{tab:duke}. Due to the challenging nature of the DukeMTMC-reID dataset, traditional re-id methods such as ResNet50 and MobileNet have obtained poor performance. Furthermore, we do not observe a good performance from the MLFN model, despite being a re-id-specific model. Irrespective of the model structure and its original performance (i.e results without auxiliary information), we observe a substantial performance gain when the clothing information is added. For instance, this gain is more than 10\% for the best performing model (OsNet) when considering the mAP metric. Moreover, the Age/Gender information has further improved this accuracy gain. These results clearly verify our hypothesis that the selected auxiliary information offer salient information cues to support the person identification. In addition, with the DukeMTMC-reID dataset, we observe the contribution that the attention-based feature fusion scheme provides to identify important features across the combined feature representation. We observe a performance gain across all the considered models when a self-attention block is added after the concatenation of main-stream features together with the auxiliary features. This elaborates that with sufficient training data, the models can further refine the combined features and achieve more robust results. 
\begin{table*}[htbp]
\resizebox{\textwidth}{!}{
\begin{tabular}{|c|c|c|c|c|c|c|c|c|c|c|c|c|c|}
\hline
                                                   & Metric  & ResNet50 & ResNet50-Concat & ResNet50-Att & MobileNet & MobileNet-Concat & MobileNet-Att & OsNet  & OsNet-Concat & OsNet-Att & MLFN   & MLFN-Concat & MLFN-Att \\ \hline
\multirow{5}{*}{DUKEMTMC—reID}                      & mAP     & 3.1\%    & -               & -            & 1.3\%     & -                & -             & 71.1\% & -            & -         & 21.1\% & -           & -        \\ \cline{2-14} 
                                                   & Rank-1  & 2.1\%    & -               & -            & 0.6\%     & -                & -             & 74.6\% & -            & -         & 18.5\% & -           & -        \\ \cline{2-14} 
                                                   & Rank-5  & 6.6\%    & -               & -            & 2.4\%     & -                & -             & 86.6\% & -            & -         & 37.7\% & -           & -        \\ \cline{2-14} 
                                                   & Rank-10 & 9.7\%    & -               & -            & 3.7\%     & -                & -             & 89.0\% & -            & -         & 49.9\% & -           & -        \\ \cline{2-14} 
                                                   & Rank-20 & 14.7\%   & -               & -            & 5.3\%     & -                & -             & 89.9\% & -            & -         & 60.7\% & -           & -        \\ \hline
\multirow{5}{*}{DUKEMTMC—reID + Clothing}              & mAP     & -        & 22.9\%          & 23.9\%       & -         & 2.1\%            & 2.2\%         & -      & 82.5\%       & 84.4\%    & -      & 25.0\%      & 26.0\%   \\ \cline{2-14} 
                                                   & Rank-1  & -        & 22.2\%          & 22.4\%       & -         & 0.9\%            & 2.0\%         & -      & 84.2\%       & 86.5\%    & -      & 24.4\%      & 23.8\%   \\ \cline{2-14} 
                                                   & Rank-5  & -        & 41.6\%          & 43.2\%       & -         & 5.0\%            & 5.1\%         & -      & 95.6\%       & 96.0\%    & -      & 45.6\%      & 47.9\%   \\ \cline{2-14} 
                                                   & Rank-10 & -        & 52.6\%          & 53.3\%       & -         & 7.8\%            & 7.1\%         & -      & 97.3\%       & 97.4\%    & -      & 55.0\%      & 56.7\%   \\ \cline{2-14} 
                                                   & Rank-20 & -        & 63.5\%          & 64.4\%       & -         & 13.7\%           & 10.4\%        & -      & 98.4\%       & 98.0\%    & -      & 67.0\%      & 68.2\%   \\ \hline
\multirow{5}{*}{DUKEMTMC—reID + Clothing + Age/Gender} & mAP     & -        & 24.6\%          & 25.2\%       & -         & 4.2\%            & 3.1\%         & -      & 87.6\%       & 88.1\%    & -      & 35.2\%      & 36.3\%   \\ \cline{2-14} 
                                                   & Rank-1  & -        & 23.1\%          & 24.5\%       & -         & 3.0\%            & 1.7\%         & -      & 89.6\%       & 89.9\%    & -      & 35.2\%      & 35.2\%   \\ \cline{2-14} 
                                                   & Rank-5  & -        & 44.3\%          & 46.1\%       & -         & 9.0\%            & 6.4\%         & -      & 96.7\%       & 96.4\%    & -      & 57.1\%      & 59.1\%   \\ \cline{2-14} 
                                                   & Rank-10 & -        & 56.1\%          & 56.3\%       & -         & 13.2\%           & 11.4\%        & -      & 98.0\%       & 97.6\%    & -      & 65.8\%      & 68.9\%   \\ \cline{2-14} 
                                                   & Rank-20 & -        & 67.7\%          & 67.7\%       & -         & 19.2\%           & 18.1\%        & -      & 98.3\%       & 98.3\%    & -      & 76.4\%      & 77.6\%   \\ \hline
\end{tabular}}
\vspace{-5mm}
\caption{Evaluation Results for the Duke Multi-Tracking Multi-Camera ReIDentification (DukeMTMC-reID) \cite{ristani2016performance} dataset.}
\label{tab:duke}
\end{table*}

\subsubsection{Results for the MARS dataset}
The evaluations of the MARS dataset with respect to the performance of ResNet50, MobileNet, OsNet, and MLFN models and their concatenation and attention-based feature fusion variants are presented in Tab. \ref{tab:mars}. Similar to the previous evaluations we observe superior results in the OsNet model and consistent contribution from both clothing and age/gender information. Furthermore, similar to the DukeMTMC-reID dataset, in the MARS dataset, we also observe a consistent performance increase across all the models when comparing the attention-based feature fusion strategy together with the naive concatenation. Both DukeMTMC-reID and MARS datasets are large-scale datasets compared to PRID20111, as such the models have access to sufficient training data to train the additional parameters of the attention layer. Therefore, we are able to see the practical utility of the attention scheme. Furthermore, with the evaluations of the MARS dataset, we observe that the ResNet-50 model that receives clothing type and Age/Gender auxiliary information has been able to achieve person-reid performance which is sufficiently closer to the state-of-the-art OsNet model. Despite the fact that the ResNet-50 model does not incorporate any architectural or training components that are specifically designed for person re-id. We believe this is a crucial observation, which allows us to clearly demonstrate the need to reconsider the information cues that are used as inputs in person re-id frameworks.

\begin{table*}[htbp]
\resizebox{\textwidth}{!}{
\begin{tabular}{|c|c|c|c|c|c|c|c|c|c|c|c|c|c|}
\hline
                                           & Metric  & ResNet50 & ResNet50-Concat & ResNet50-Att & MobileNet & MobileNet-Concat & MobileNet-Att & OsNet  & OsNet-Concat & OsNet-Att & MLFN   & MLFN-Concat & MLFN-Att \\ \hline
\multirow{5}{*}{MARS}                      & mAP     & 34.2\%   & -               & -            & 30.7\%    & -                & -             & 53.8\% & -            & -         & 20.3\% & -           & -        \\ \cline{2-14} 
                                           & Rank-1  & 47.9\%   & -               & -            & 43.6\%    & -                & -             & 70.3\% & -            & -         & 30.1\% & -           & -        \\ \cline{2-14} 
                                           & Rank-5  & 65.2\%   & -               & -            & 62.0\%    & -                & -             & 82.5\% & -            & -         & 48.3\% & -           & -        \\ \cline{2-14} 
                                           & Rank-10 & 72.6\%   & -               & -            & 69.6\%    & -                & -             & 85.4\% & -            & -         & 55.8\% & -           & -        \\ \cline{2-14} 
                                           & Rank-20 & 78.5\%   & -               & -            & 76.2\%    & -                & -             & 87.8\% & -            & -         & 63.9\% & -           & -        \\ \hline
\multirow{5}{*}{MARS+ Clothing}               & mAP     & -        & 43.0\%          & 49.7\%       & -         & 32.5\%           & 33.4\%        & -      & 58.9\%       & 61.0\%    & -      & 25.8\%      & 29.5\%   \\ \cline{2-14} 
                                           & Rank-1  & -        & 55.7\%          & 62.8\%       & -         & 45.4\%           & 46.9\%        & -      & 76.2\%       & 77.7\%    & -      & 37.7\%      & 41.0\%   \\ \cline{2-14} 
                                           & Rank-5  & -        & 72.0\%          & 78.8\%       & -         & 63.6\%           & 65.8\%        & -      & 88.0\%       & 88.7\%    & -      & 54.8\%      & 59.8\%   \\ \cline{2-14} 
                                           & Rank-10 & -        & 77.8\%          & 83.2\%       & -         & 71.0\%           & 72.8\%        & -      & 90.2\%       & 90.8\%    & -      & 62.8\%      & 67.4\%   \\ \cline{2-14} 
                                           & Rank-20 & -        & 82.2\%          & 86.7\%       & -         & 77.3\%           & 80.1\%        & -      & 91.8\%       & 92.4\%    & -      & 70.6\%      & 73.8\%   \\ \hline
\multirow{5}{*}{MARS + Clothing + Age/Gender} & mAP     & -        & 51.9\%          & 52.3\%       & -         & 34.0\%           & 36.3\%        & -      & 59.5\%       & 62.3\%    & -      & 34.9\%      & 37.9\%   \\ \cline{2-14} 
                                           & Rank-1  & -        & 66.1\%          & 69.6\%       & -         & 47.3\%           & 49.0\%        & -      & 75.5\%       & 77.0\%    & -      & 47.8\%      & 49.3\%   \\ \cline{2-14} 
                                           & Rank-5  & -        & 80.9\%          & 83.0\%       & -         & 66.3\%           & 68.2\%        & -      & 88.0\%       & 89.1\%    & -      & 66.5\%      & 70.8\%   \\ \cline{2-14} 
                                           & Rank-10 & -        & 85.3\%          & 85.9\%       & -         & 74.7\%           & 75.2\%        & -      & 90.2\%       & 91.6\%    & -      & 74.0\%      & 77.3\%   \\ \cline{2-14} 
                                           & Rank-20 & -        & 88.9\%          & 88.2\%       & -         & 80.7\%           & 81.4\%        & -      & 91.8\%       & 93.2\%    & -      & 79.2\%      & 82.9\%   \\ \hline
\end{tabular}}
\vspace{-5mm}
\caption{Evaluation results for the Motion Analysis and Re-identification Set (MARS) \cite{zheng2016mars} dataset.}
\label{tab:mars}
\end{table*}

\subsubsection{Results for the MSU-AVIS dataset}
The evaluations of the ResNet50, MobileNet, OsNet, and MLFN models and their concatenation and attention-based fusion variants on the MSU-AVIS dataset are presented in Tab. \ref{tab:msuavis}. When considering the presented results we observe substantial performance gain with the introduction of clothing and audio information. For instance, the state-of-the-art OsNet model achieves an mAP of 100\% when the clothing information is introduced with attention-based feature fusion. When considering the attention augmented feature fusion and naive feature concatenation, similar to previous evaluations, we observe significant robustness when the features are fused using an attention-based mechanism.

\begin{table*}[htbp]
\resizebox{\textwidth}{!}{
\begin{tabular}{|c|c|c|c|c|c|c|c|c|c|c|c|c|c|}
\hline
                                                            & Metric & ResNet50 & ResNet50-Concat & ResNet50-Att & MobileNet & MobileNet-Concat & MobileNet-Att & OsNet  & OsNet-Concat & OsNet-Att & MLFN   & MLFN-Concat & MLFN-Att \\ \hline
\multirow{3}{*}{MSU-AVIS}                                   & mAP    & 53.9\%   & -               & -            & 59.5\%    & -                & -             & 61.0\% &              & -         & 53.9\% &             & -        \\ \cline{2-14} 
                                                            & Rank-1 & 71.8\%   & -               & -            & 75.5\%    & -                & -             & 77.7\% &              & -         & 71.8\% &             & -        \\ \cline{2-14} 
                                                            & Rank-5 & 82.9\%   & -               & -            & 88.0\%    & -                & -             & 88.7\% &              & -         & 82.9\% &             & -        \\ \hline
\multirow{3}{*}{MSU-AVIS+Clothing}                            & mAP    & -        & 86.7\%          & 91.4\%       & -         & 90.0\%           & 96.0\%        & -      & 97.3\%       & 100.0\%   & -      & 57.5\%      & 68.7\%   \\ \cline{2-14} 
                                                            & Rank-1 & -        & 76.0\%          & 88.0\%       & -         & 80.0\%           & 92.0\%        & -      & 96.0\%       & 100.0\%   & -      & 48.0\%      & 56.0\%   \\ \cline{2-14} 
                                                            & Rank-5 & -        & 100.0\%         & 96.0\%       & -         & 100.0\%          & 100.0\%       & -      & 100.0\%      & 100.0\%   & -      & 72.0\%      & 88.0\%   \\ \hline
\multirow{3}{*}{MSU-AVIS+Clothing + Age/Gender + Audio}       & mAP    & -        & 95.3\%          & 98.0\%       & -         & 95.3\%           & 100.0\%       & -      & 98.0\%       & 100.0\%   & -      & 75.5\%      & 87.6\%   \\ \cline{2-14} 
                                                            & Rank-1 & -        & 92.0\%          & 96.0\%       & -         & 92.0\%           & 100.0\%       & -      & 96.0\%       & 100.0\%   & -      & 68.0\%      & 80.0\%   \\ \cline{2-14} 
                                                            & Rank-5 & -        & 100.0\%         & 100.0\%      & -         & 100.0\%          & 100.0\%       & -      & 100.0\%      & 100.0\%   & -      & 88.0\%      & 96.0\%   \\ \hline
\multirow{3}{*}{MSU-AVIS + Age/Gender + Audio + Trajectory} & mAP    & -        & 94.6\%          & 98.0\%       & -         & 98.0\%           & 100.0\%       & -      & 100.0\%      & 100.0\%   & -      & 84.7\%      & 96.0\%   \\ \cline{2-14} 
                                                            & Rank-1 & -        & 92.0\%          & 96.0\%       & -         & 96.0\%           & 100.0\%       & -      & 100.0\%      & 100.0\%   & -      & 76.0\%      & 100.0\%  \\ \cline{2-14} 
                                                            & Rank-5 & -        & 96.0\%          & 100.0\%      & -         & 100.0\%          & 100.0\%       & -      & 100.0\%      & 100.0\%   & -      & 100.0\%     & 100.0\%  \\ \hline
\end{tabular}}
\vspace{-5mm}
\caption{Evaluation results of the MSU-AVIS dataset \cite{chowdhury2018msu}. }
\label{tab:msuavis}
\end{table*}

\subsubsection{Results for the NBA Players dataset}

In our next evaluation, we specifically evaluated the contribution from the tattoo information using the NBA Players dataset and the results are presented in Tab. \ref{tab:nbaplayers}. The poor result that the baseline models achieve clearly demonstrates the challenges associated with this dataset. When analysing the presented results we observe several interesting characteristics.

First, in contrast to the previous evaluations, we observe only a slight performance increase with the addition of clothing information. We believe this is due to the fact that athletes in our dataset can wear similar jerseys which can lead to confusion. However, a significant boost in performance is observed with the addition of the tattoo information, which have led to the extraction of unique traits to identify player identities. Second, similar to the MARS dataset, both ResNet50 and OsNet models have achieved highly comparable results, especially in the evaluation settings NBA\_Players+ Tattoos+Cloths+ Age/Gender, NBA\_Players+ Tattoos, and NBA\_Players+ Tattoos+Cloths. In the last two evaluation settings, the ResNet50 model has even surpassed the OsNet, which clearly illustrates the contributions from the information cues that the auxiliary information provide. 

\begin{table*}[htbp]
\resizebox{\textwidth}{!}{
\begin{tabular}{|c|c|c|c|c|c|c|c|c|c|c|c|c|c|}
\hline
                                                          & Metric  & ResNet50 & ResNet50-Concat & ResNet50-Att & MobileNet & MobileNet-Concat & MobileNet-Att & OsNet  & OsNet-Concat & OsNet-Att & MLFN   & MLFN-Concat & MLFN-Att \\ \hline
\multirow{5}{*}{NBA\_Players}                             & mAP     & 7.1\%    & -               & -            & 4.7\%     & -                & -             & 7.4\%  &              & -         & 4.9\%  &             & -        \\ \cline{2-14} 
                                                          & Rank-1  & 10.1\%   & -               & -            & 4.0\%     & -                & -             & 11.1\% &              & -         & 2.0\%  &             & -        \\ \cline{2-14} 
                                                          & Rank-5  & 33.3\%   & -               & -            & 22.2\%    & -                & -             & 32.3\% &              & -         & 21.2\% &             & -        \\ \cline{2-14} 
                                                          & Rank-10 & 49.5\%   & -               & -            & 33.3\%    & -                & -             & 51.5\% &              & -         & 31.3\% &             & -        \\ \cline{2-14} 
                                                          & Rank-20 & 71.7\%   & -               & -            & 52.5\%    & -                & -             & 72.7\% &              & -         & 55.6\% &             & -        \\ \hline
\multirow{5}{*}{NBA\_Players+ Clothing}                      & mAP     & -        & 7.7\%           & 10.8\%       & -         & 5.0\%            & 5.3\%         & -      & 7.9\%        & 12.4\%    & -      & 5.0\%       & 8.4\%    \\ \cline{2-14} 
                                                          & Rank-1  & -        & 13.1\%          & 16.2\%       & -         & 4.0\%            & 7.1\%         & -      & 14.1\%       & 15.2\%    & -      & 6.1\%       & 3.0\%    \\ \cline{2-14} 
                                                          & Rank-5  & -        & 38.4\%          & 29.3\%       & -         & 20.2\%           & 17.2\%        & -      & 28.3\%       & 24.2\%    & -      & 16.2\%      & 12.1\%   \\ \cline{2-14} 
                                                          & Rank-10 & -        & 50.5\%          & 36.4\%       & -         & 36.4\%           & 34.3\%        & -      & 50.5\%       & 28.3\%    & -      & 28.3\%      & 16.2\%   \\ \cline{2-14} 
                                                          & Rank-20 & -        & 69.7\%          & 48.5\%       & -         & 51.5\%           & 55.6\%        & -      & 75.8\%       & 31.3\%    & -      & 54.5\%      & 18.2\%   \\ \hline
\multirow{5}{*}{NBA\_Players+ Clothing+ Age/Gender}          & mAP     & -        & 12.0\%          & 10.8\%       & -         & 5.0\%            & 5.2\%         & -      & 8.0\%        & 12.3\%    & -      & 5.5\%       & 8.4\%    \\ \cline{2-14} 
                                                          & Rank-1  & -        & 12.1\%          & 16.2\%       & -         & 6.1\%            & 8.1\%         & -      & 10.1\%       & 11.1\%    & -      & 6.1\%       & 13.1\%   \\ \cline{2-14} 
                                                          & Rank-5  & -        & 27.3\%          & 29.3\%       & -         & 21.2\%           & 23.2\%        & -      & 35.4\%       & 22.2\%    & -      & 21.2\%      & 32.3\%   \\ \cline{2-14} 
                                                          & Rank-10 & -        & 39.4\%          & 36.4\%       & -         & 31.3\%           & 36.4\%        & -      & 50.5\%       & 26.3\%    & -      & 33.3\%      & 43.4\%   \\ \cline{2-14} 
                                                          & Rank-20 & -        & 56.6\%          & 48.5\%       & -         & 46.5\%           & 54.5\%        & -      & 79.8\%       & 34.3\%    & -      & 57.6\%      & 62.6\%   \\ \hline
\multirow{5}{*}{NBA\_Players+ Tattoos}                    & mAP     & -        & 11.3\%          & 17.0\%       & -         & 5.0\%            & 5.0\%         & -      & 11.2\%       & 15.6\%    & -      & 5.2\%       & 10.3\%   \\ \cline{2-14} 
                                                          & Rank-1  & -        & 18.2\%          & 29.3\%       & -         & 5.1\%            & 7.1\%         & -      & 18.2\%       & 38.4\%    & -      & 6.1\%       & 14.1\%   \\ \cline{2-14} 
                                                          & Rank-5  & -        & 55.6\%          & 62.6\%       & -         & 17.2\%           & 21.2\%        & -      & 47.5\%       & 60.6\%    & -      & 16.2\%      & 36.4\%   \\ \cline{2-14} 
                                                          & Rank-10 & -        & 63.6\%          & 76.8\%       & -         & 31.3\%           & 29.3\%        & -      & 69.7\%       & 74.7\%    & -      & 34.3\%      & 48.5\%   \\ \cline{2-14} 
                                                          & Rank-20 & -        & 72.7\%          & 88.9\%       & -         & 54.5\%           & 54.5\%        & -      & 81.8\%       & 87.9\%    & -      & 54.5\%      & 70.7\%   \\ \hline
\multirow{5}{*}{NBA\_Players+ Tattoos+Clothing}             & mAP     & -        &                 & 17.5\%       & -         & -                & 5.3\%         & -      & -            & 10.5\%    & -      & -           & 8.6\%    \\ \cline{2-14} 
                                                          & Rank-1  & -        &                 & 26.3\%       & -         & -                & 10.1\%        & -      & -            & 24.2\%    & -      & -           & 10.1\%   \\ \cline{2-14} 
                                                          & Rank-5  & -        &                 & 54.5\%       & -         & -                & 24.2\%        & -      & -            & 45.5\%    & -      & -           & 33.3\%   \\ \cline{2-14} 
                                                          & Rank-10 & -        &                 & 66.7\%       & -         & -                & 36.4\%        & -      & -            & 50.5\%    & -      & -           & 45.5\%   \\ \cline{2-14} 
                                                          & Rank-20 & -        &                 & 72.7\%       & -         & -                & 55.6\%        & -      & -            & 57.6\%    & -      & -           & 54.5\%   \\ \hline
\multirow{5}{*}{NBA\_Players+ Tattoos+Clothing+ Age/Gender} & mAP     & -        & 7.0\%           & 17.6\%       & -         & 5.0\%            & 6.5\%         & -      & 10.6\%       & 18.5\%    & -      & 5.3\%       & 8.9\%    \\ \cline{2-14} 
                                                          & Rank-1  & -        & 12.1\%          & 30.3\%       & -         & 7.1\%            & 7.1\%         & -      & 13.1\%       & 39.4\%    & -      & 6.1\%       & 10.1\%   \\ \cline{2-14} 
                                                          & Rank-5  & -        & 29.3\%          & 65.7\%       & -         & 16.2\%           & 20.2\%        & -      & 19.2\%       & 69.7\%    & -      & 18.2\%      & 32.3\%   \\ \cline{2-14} 
                                                          & Rank-10 & -        & 48.5\%          & 79.8\%       & -         & 30.3\%           & 38.4\%        & -      & 22.2\%       & 81.8\%    & -      & 35.4\%      & 47.5\%   \\ \cline{2-14} 
                                                          & Rank-20 & -        & 67.7\%          & 91.9\%       & -         & 52.5\%           & 65.7\%        & -      & 30.3\%       & 91.9\%    & -      & 53.5\%      & 62.6\%   \\ \hline
\multirow{5}{*}{NBA\_Players+ Tattoos+Logos+ Age/Gender}  & mAP     & -        & 9.9\%           & 14.7\%       & -         & 5.0\%            & 5.2\%         & -      & 10.5\%       & 21.3\%    & -      & 6.6\%       & 8.6\%    \\ \cline{2-14} 
                                                          & Rank-1  & -        & 16.2\%          & 29.3\%       & -         & 5.1\%            & 6.1\%         & -      & 20.2\%       & 45.5\%    & -      & 9.1\%       & 15.2\%   \\ \cline{2-14} 
                                                          & Rank-5  & -        & 41.4\%          & 64.6\%       & -         & 21.2\%           & 22.2\%        & -      & 43.4\%       & 73.7\%    & -      & 19.2\%      & 37.4\%   \\ \cline{2-14} 
                                                          & Rank-10 & -        & 53.5\%          & 73.7\%       & -         & 37.4\%           & 31.3\%        & -      & 64.6\%       & 81.8\%    & -      & 30.3\%      & 52.5\%   \\ \cline{2-14} 
                                                          & Rank-20 & -        & 68.7\%          & 88.9\%       & -         & 50.5\%           & 50.5\%        & -      & 77.8\%       & 89.9\%    & -      & 44.4\%      & 72.7\%   \\ \hline
\end{tabular}}
\vspace{-5mm}
\caption{Evaluation results for the NBA Players dataset.}
\label{tab:nbaplayers}
\end{table*}

\subsubsection{Qualitative Results}

\begin{figure*}[htbp]
    \centering
    \includegraphics[width= \textwidth]{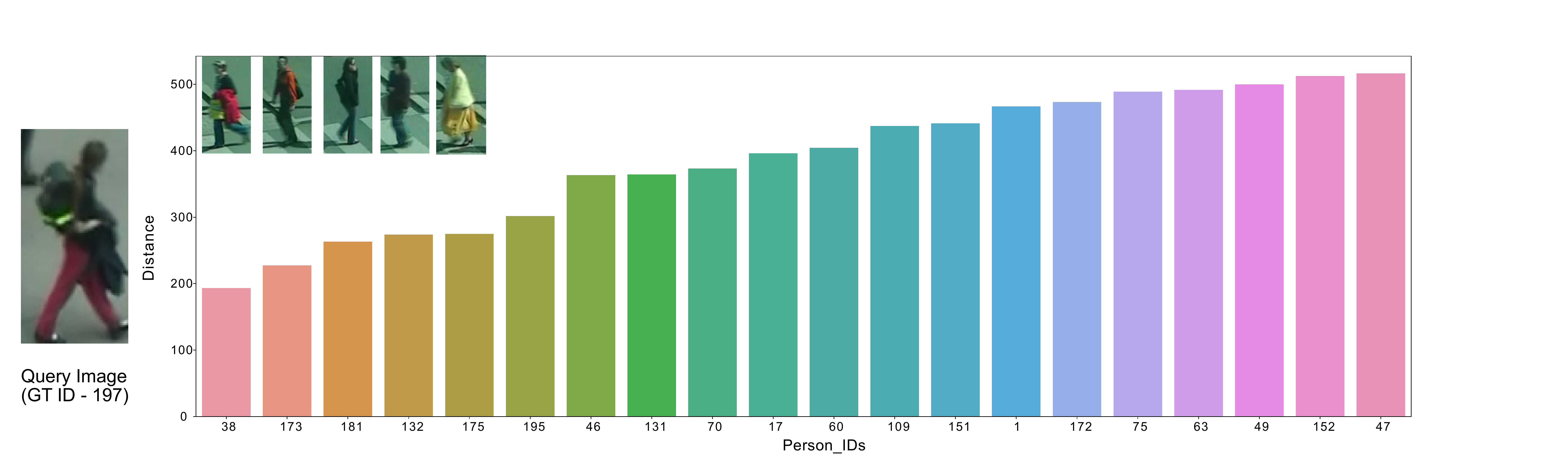}
    \caption{Qualitative results of the baseline OSNet model for a sample test image from PRID2011 test set.}
    \label{fig:baseline_qualitative}
\end{figure*}

\begin{figure*}[htbp]
    \centering
    \includegraphics[width= \textwidth]{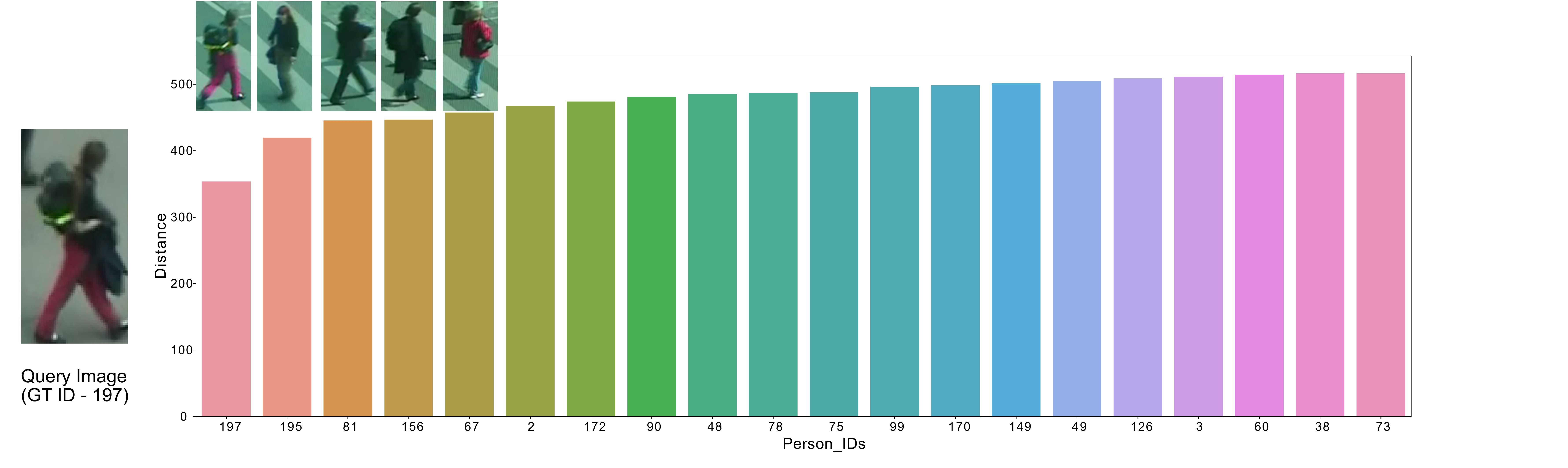}
    \caption{Qualitative results of the OSNet-Att model that receives auxiliary clothing type and age/gender information for a sample test image from PRID2011 test set.}
    \label{fig:proposed_qualitative}
\end{figure*}

Qualitative results of an interesting test example are presented in Figs. \ref{fig:baseline_qualitative} and \ref{fig:proposed_qualitative}. We selected an example from the PRID2011 test set and conducted this comparison using the baseline OSNet model and OSNet-Att model that receives auxiliary clothing type and age/gender information. Note that these experiments were conducted only using the OSNet model as it is the best performing model. In Fig. \ref{fig:baseline_qualitative}, we present the feature distances resultant from the baseline OSNet model for the top-20 ranked images in the gallery set for the query image (shown on the left-hand side). We observe that subjects with red/maroon colour clothing have obtained smaller distances irrespective of their clothing types. In contrast, when considering the outputs of the OSNet-Att model with auxiliary clothing type and age/gender information presented in \ref{fig:proposed_qualitative}, we observe that such erroneous matchings have been removed. We believe this ability of the model owes to the utilisation of the auxiliary clothing information which allows it to compare and contrast different clothing types, and age/gender characteristics and learn a more informed feature representation.

\subsubsection{Model Interpretation Outputs}

To understand how the person re-identification method evaluates the additional information that we combine, we utilise the Integrated Gradient (IG) \cite{sundararajan2017axiomatic} model interpretation mechanism. This method calculates the gradient of the model's prediction to its input features. As such we can visualise the input feature importance that contributes to the model's accurate predictions. This experiment is conducted using our NBA Players dataset and we use the classifier of our person re-identification method \footnote{For the ease of the computations we only evaluate this for the OsNet-Att model, which is the best performing model.} and compute the gradients of the classifier predictions with respect to its inputs (i.e original features of the model and the auxiliary features that we input). When computing the IGs we randomly selected 100 test queries and average the IG attributions.

\begin{figure*}[htbp]
     \centering
     \begin{subfigure}[b]{1\textwidth}
         \centering
         \includegraphics[width=\textwidth]{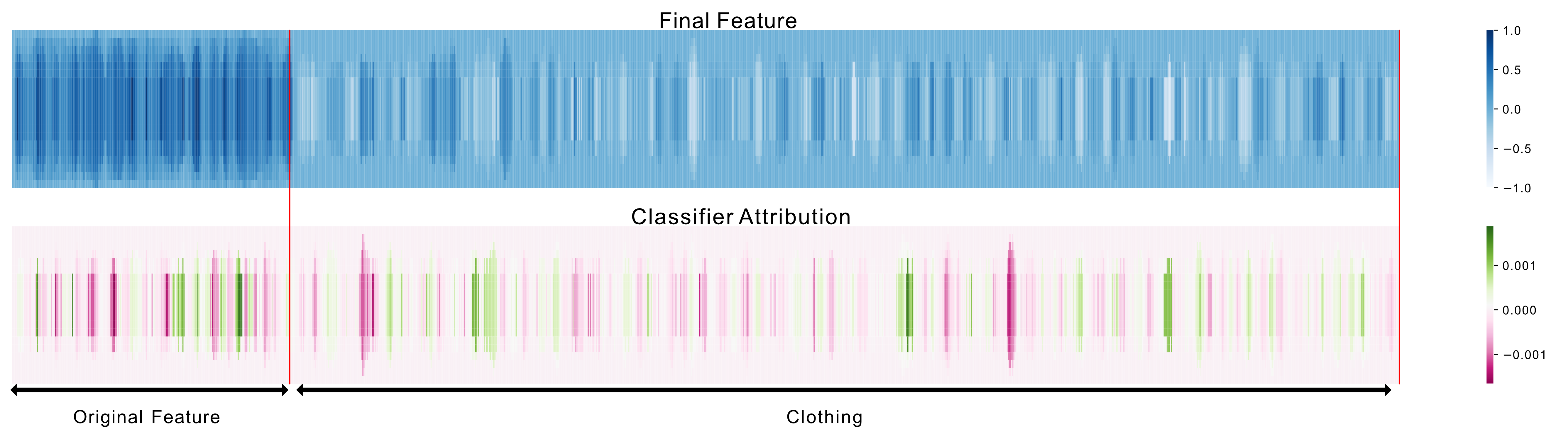}
         \caption{With Clothing information.}
     \end{subfigure}
     \hfill
     \begin{subfigure}[b]{1\textwidth}
         \centering  
         \includegraphics[width=\textwidth]{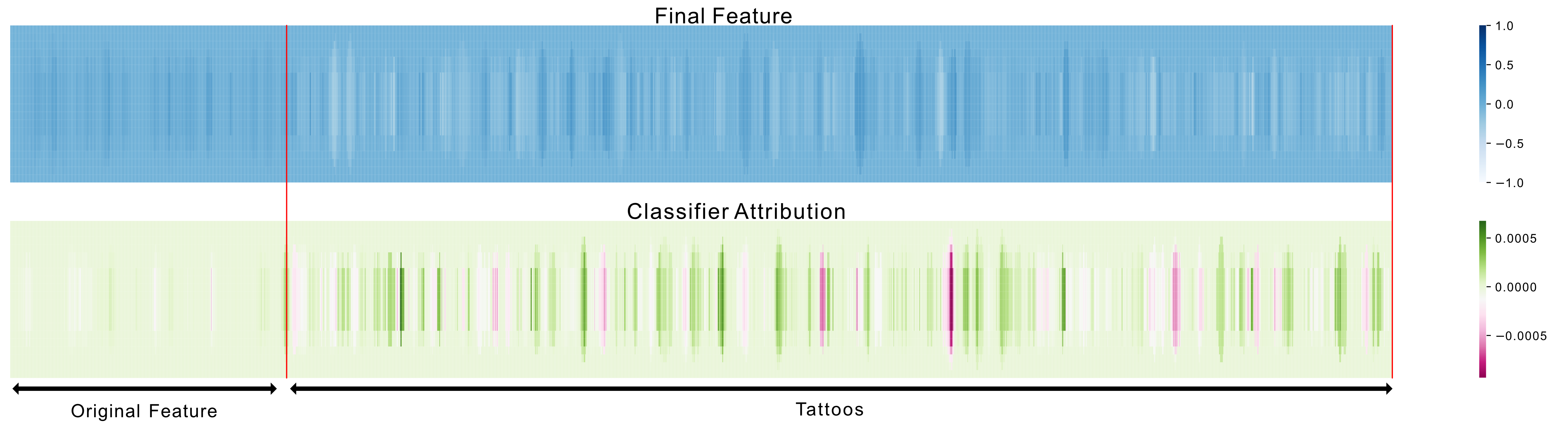}
         \caption{With Tattoos information.}
     \end{subfigure}
     \hfill
     \begin{subfigure}[b]{1\textwidth}
         \centering
         \includegraphics[width=\textwidth]{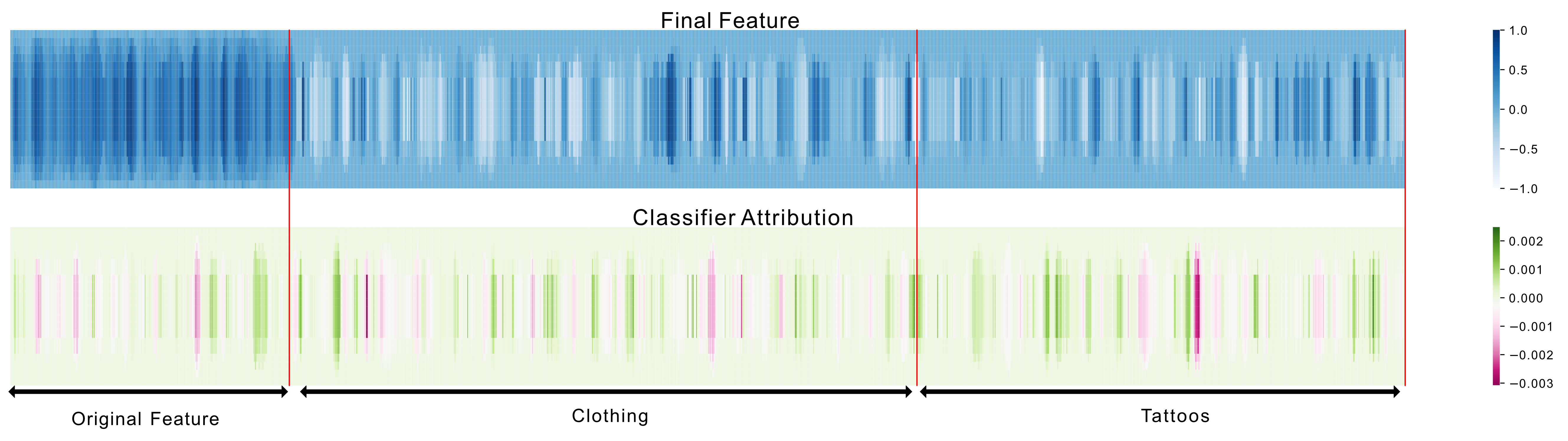}
         \caption{With Clothing and Tattoos information.}
     \end{subfigure}
     \caption{Person Re-id classifier attributions generated using the Integrated Gradient (IG) model interpretation method. }
          \label{fig:attributions}

\end{figure*} 

When analysing the presented attribution results in Fig. \ref{fig:attributions}, we observe interesting behaviour in our model in the presence of different auxiliary feature combinations. For instance, when considering the positive attributions in Fig. \ref{fig:attributions} (a) we observe that the model attends to both original features (i.e. features from the original OsNet model) and the clothing information equally. In contrast, in Fig. \ref{fig:attributions} (b), we see a majority of attributions from the tattoo feature, which clearly illustrates the superiority of this information compared to the original features from the OsNet model. Furthermore, in Fig. \ref{fig:attributions} (c), we see attributions from all the information sources, indicating that the model selects the necessary information from all the feature types and is not completely driven by the tattoo information.

\section{Limitations and Open Research Challenges}\label{sec:limitations}

In this paper, we have investigated the utility of auxiliary information sources such as clothing type, voice, tattoos, logos, and behaviour-based trajectories for person re-identification. Our evaluations suggest a substantial performance gain with the introduction of these information sources, irrespective of the model design. For instance, the auxiliary information helped the state-of-the-art OSNet model which captures features from a set of heterogeneous scale representations, extracting both fine-grained attributes from a person of interest and the feature combinations, to improve its performance. In addition, a simpler off-the-shelf ResNet 50 model was also able to improve its re-id performance, while in some cases achieving competitive results with respect to OSNet. As such, there is a clear utility of incorporating the auxiliary into the feature descriptor and our interpretation outputs describe how it helps the model to learn a more discriminative identity representation which otherwise fails. Consequently, we are able to introduce a novel paradigm in person re-id where explicit auxiliary attribute representations are fused with the person re-id model features, rendering an augmented feature descriptor. To the best of our knowledge, this is the first work to investigate such a concept for person re-id.

For future works, we have identified four key research areas. First, the selected set of auxiliary information is not an exhaustive set. Attributes such as objects they carry, the subject's footwear type, and other wearables can be detected and features from those attributes can also be incorporated into the re-id framework. This can yield better results as it further improves the discriminative power of the rendered feature descriptor. Moreover, this framework can be further extended to consider background context in addition to the identity-specific cues to support image search. For instance, in a law enforcement-type application, the end users can utilise cues in the image background and recognise geographical locations as well as suspicious behavioural patterns based on such background information. Therefore, it is particularly interesting to incorporate such descriptors in a dynamic fusion framework that can pay varying levels of attention to the auxiliary features depending on the application. 

Another interesting research direction is the development of methods to mitigate the domain shifts within the extracted same auxiliary feature from different views as well as the domain discrepancies that arise due to the density differences of different auxiliary feature types. Similar to the approaches proposed in cross-domain person re-id literature, one possible avenue to explore is the utilisation of cycle-GAN-based domain translation strategies to learn a domain invariant feature representation. In addition to learning such invariant descriptors, the GAN will also learn to synthesise auxiliary features in a new camera view, which can be valuable in scenarios such as occluded person attributes and poor image resolution. Moreover, GANs can also be utilised to translate between auxiliary features and learn an invariant representation. 

As the third research direction better feature fusion strategies and model learning strategies can be identified. For instance, our experimental evaluations suggested the importance of attention-based feature selection in fusion. However, we observed that these trainable parameters could affect the model training process in a setting where there are limited training data. As such more robust feature fusion strategies can be investigated where there are training data scarcities. For instance, soft and hard-wired attention \cite{fernando2018soft} based feature fusion strategy can be devised to incorporate pre-defined heuristic into the feature fusion framework. In addition, where there is sufficient training data, more complex, yet robust fusion strategies such as graph-based feature fusion \cite{priyasad2022affect, liao2013graph} or trainable neural memory networks based feature fusion \cite{priyasad2021memory, gammulle2019forecasting} can be leveraged. For the simplicity of our evaluations, we did not utilised complex model training strategies in our experiments. Therefore, in future works more sophisticated losses such as dense triplet loss can be incorporated into the training phase. In addition, novel losses can be devised using knowledge distillation loss or identity consistency loss with the aim of mitigating the domain shifts that could arise when learning with diverse auxiliary features. 

Finally, we would also like to indulge the reader to explore mechanisms to reduce the computational overhead that occurs when evaluating multiple auxiliary feature extractors and when combining large auxiliary feature descriptors. Lightweight detector backbones should be investigated for the image-based feature descriptors. Similarly, model pruning strategies can be leveraged to alleviate the computational overhead in audio and other information extractors and reduced to the dimensionality of the synthesised descriptor.

\section{Conclusion}\label{sec:conclusion}

In this paper, we discussed a novel approach for improving the performance of existing person re-identification (re-id) techniques using auxiliary feature information. In particular, we provided a structured analysis of the existing state-of-the-art methods in both uni-modal and multi-modal person re-id, discussing their limitations and elaborated on how these methods can be extended to support learning with auxiliary attributes. To the best of our knowledge, such information fusion strategies have never been exploited in person re-id and as such, we introduce a novel paradigm in the person re-id domain. Furthermore, we perform a quantitative analysis of the proposed hypothesis using both popular person re-id models and popular image-based models on four public person re-id benchmarks and one dataset that we have collected. Investigations have been conducted using both concatenation and attention-based fusion of the extracted auxiliary features with the primary features from the re-id model. With the generated model interpretation outputs we were able to confirm our intuition that auxiliary features are useful for identifying subjects in challenging situations.  As concluding remarks, we outlined key limitations of our techniques and propose interesting future research directions for further investigation.

\section{Acknowledgement}
The research presented in this paper was supported by the Australian Research Council (ARC) Discovery grant DP200101942 and by the Artificial Intelligence for Decision Making (AI4DM) research initiative grant (No: 11432) from the Defence Science \& Technology Group of the Department of Defence.

\bibliography{refs}

\begin{thebibliography}{10}
\expandafter\ifx\csname url\endcsname\relax
  \def\url#1{\texttt{#1}}\fi
\expandafter\ifx\csname urlprefix\endcsname\relax\def\urlprefix{URL }\fi
\expandafter\ifx\csname href\endcsname\relax
  \def\href#1#2{#2} \def\path#1{#1}\fi

\bibitem{wu2018exploit}
Y.~Wu, Y.~Lin, X.~Dong, Y.~Yan, W.~Ouyang, Y.~Yang, Exploit the unknown
  gradually: One-shot video-based person re-identification by stepwise
  learning, in: Computer vision and pattern recognition, 2018.

\bibitem{zheng2019re}
M.~Zheng, S.~Karanam, Z.~Wu, R.~J. Radke, Re-identification with consistent
  attentive siamese networks, in: Computer vision and pattern recognition,
  2019, pp. 5735--5744.

\bibitem{tay2019aanet}
C.-P. Tay, S.~Roy, K.-H. Yap, Aanet: Attribute attention network for person
  re-identifications, in: Conference on computer vision and pattern
  recognition, 2019, pp. 7134--7143.

\bibitem{liao2018video}
X.~Liao, L.~He, Z.~Yang, C.~Zhang, Video-based person re-identification via 3d
  convolutional networks and non-local attention, in: Asian Conference on
  Computer Vision, Springer, 2018, pp. 620--634.

\bibitem{chung2017two}
D.~Chung, K.~Tahboub, E.~J. Delp, A two stream siamese convolutional neural
  network for person re-identification, in: Conference on computer vision,
  2017, pp. 1983--1991.

\bibitem{fu2019sta}
Y.~Fu, X.~Wang, Y.~Wei, T.~Huang, Sta: Spatial-temporal attention for
  large-scale video-based person re-identification, in: Conference on
  artificial intelligence, Vol.~33, 2019, pp. 8287--8294.

\bibitem{huang2020real}
Y.~Huang, Z.-J. Zha, X.~Fu, R.~Hong, L.~Li, Real-world person re-identification
  via degradation invariance learning, in: Conference on Computer Vision and
  Pattern Recognition, 2020, pp. 14084--14094.

\bibitem{chen2019instance}
Y.~Chen, X.~Zhu, S.~Gong, Instance-guided context rendering for cross-domain
  person re-identification, in: International Conference on Computer Vision,
  2019, pp. 232--242.

\bibitem{liu2021spatial}
J.~Liu, Z.-J. Zha, W.~Wu, K.~Zheng, Q.~Sun, Spatial-temporal correlation and
  topology learning for person re-identification in videos, in: Computer Vision
  and Pattern Recognition, 2021, pp. 4370--4379.

\bibitem{wu2021discover}
Q.~Wu, P.~Dai, J.~Chen, C.-W. Lin, Y.~Wu, F.~Huang, B.~Zhong, R.~Ji, Discover
  cross-modality nuances for visible-infrared person re-identification, in:
  Conference on Computer Vision and Pattern Recognition, 2021.

\bibitem{wang2020cross}
G.-A. Wang, T.~Zhang, Y.~Yang, J.~Cheng, J.~Chang, X.~Liang, Z.-G. Hou,
  Cross-modality paired-images generation for rgb-infrared person
  re-identification, in: Conference on Artificial Intelligence, Vol.~34, 2020.

\bibitem{ming2022deep}
Z.~Ming, M.~Zhu, X.~Wang, J.~Zhu, J.~Cheng, C.~Gao, Y.~Yang, X.~Wei, Deep
  learning-based person re-identification methods: A survey and outlook of
  recent works, Image and Vision Computing 119 (2022) 104394.

\bibitem{ristani2016performance}
E.~Ristani, F.~Solera, R.~Zou, R.~Cucchiara, C.~Tomasi, Performance measures
  and a data set for multi-target, multi-camera tracking, in: European
  conference on computer vision, Springer, 2016, pp. 17--35.

\bibitem{li2014deepreid}
W.~Li, R.~Zhao, T.~Xiao, X.~Wang, Deepreid: Deep filter pairing neural network
  for person re-identification, in: Conference on computer vision and pattern
  recognition, 2014, pp. 152--159.

\bibitem{zhou2019omni}
K.~Zhou, Y.~Yang, A.~Cavallaro, T.~Xiang, Omni-scale feature learning for
  person re-identification, in: International Conference on Computer Vision,
  2019, pp. 3702--3712.

\bibitem{guo2018efficient}
Y.~Guo, N.-M. Cheung, Efficient and deep person re-identification using
  multi-level similarity, in: Conference on computer vision and pattern
  recognition, 2018, pp. 2335--2344.

\bibitem{ahmed2015improved}
E.~Ahmed, M.~Jones, T.~K. Marks, An improved deep learning architecture for
  person re-identification, in: Conference on computer vision and pattern
  recognition, 2015, pp. 3908--3916.

\bibitem{wang2018person}
Y.~Wang, Z.~Chen, F.~Wu, G.~Wang, Person re-identification with cascaded
  pairwise convolutions, in: Computer Vision and Pattern Recognition, 2018.

\bibitem{subramaniam2019co}
A.~Subramaniam, A.~Nambiar, A.~Mittal, Co-segmentation inspired attention
  networks for video-based person re-identification, in: International
  conference on computer vision, 2019, pp. 562--572.

\bibitem{li2018diversity}
S.~Li, S.~Bak, P.~Carr, X.~Wang, Diversity regularized spatiotemporal attention
  for video-based person re-identification, in: Conference on computer vision
  and pattern recognition, 2018, pp. 369--378.

\bibitem{wang2016person}
T.~Wang, S.~Gong, X.~Zhu, S.~Wang, Person re-identification by discriminative
  selection in video ranking, IEEE transactions on pattern analysis and machine
  intelligence 38~(12) (2016) 2501--2514.

\bibitem{li2021diverse}
Y.~Li, J.~He, T.~Zhang, X.~Liu, Y.~Zhang, F.~Wu, Diverse part discovery:
  Occluded person re-identification with part-aware transformer, in: Conference
  on Computer Vision and Pattern Recognition, 2021, pp. 2898--2907.

\bibitem{wu2021person}
S.~Wu, Y.~Bai, C.~Wang, L.~Duan, Person retrieval with conv-transformer, in:
  2021 IEEE International Conference on Multimedia and Expo (ICME), 2021.

\bibitem{chang2018multi}
X.~Chang, T.~M. Hospedales, T.~Xiang, Multi-level factorisation net for person
  re-identification, in: Conference on computer vision and pattern recognition,
  2018, pp. 2109--2118.

\bibitem{zhong2018camera}
Z.~Zhong, L.~Zheng, Z.~Zheng, S.~Li, Y.~Yang, Camera style adaptation for
  person re-identification, in: Computer vision and pattern recognition, 2018,
  pp. 5157--5166.

\bibitem{liu2019adaptive}
J.~Liu, Z.-J. Zha, D.~Chen, R.~Hong, M.~Wang, Adaptive transfer network for
  cross-domain person re-identification, in: Conference on computer vision and
  pattern recognition, 2019, pp. 7202--7211.

\bibitem{ye2018visible}
M.~Ye, Z.~Wang, X.~Lan, P.~C. Yuen, Visible thermal person re-identification
  via dual-constrained top-ranking., in: IJCAI, Vol.~1, 2018.

\bibitem{ye2018hierarchical}
M.~Ye, X.~Lan, J.~Li, P.~Yuen, Hierarchical discriminative learning for visible
  thermal person re-identification, in: Conference on Artificial Intelligence,
  Vol.~32, 2018.

\bibitem{ye2020dynamic}
M.~Ye, J.~Shen, D.~J~Crandall, L.~Shao, J.~Luo, Dynamic dual-attentive
  aggregation learning for visible-infrared person re-identification, in:
  European Conference on Computer Vision, Springer, 2020, pp. 229--247.

\bibitem{choi2020hi}
S.~Choi, S.~Lee, Y.~Kim, T.~Kim, C.~Kim, Hi-cmd: Hierarchical cross-modality
  disentanglement for visible-infrared person re-identification, in: Computer
  vision and pattern recognition, 2020, pp. 10257--10266.

\bibitem{dai2018cross}
P.~Dai, R.~Ji, H.~Wang, Q.~Wu, Y.~Huang, Cross-modality person
  re-identification with generative adversarial training., in: IJCAI, Vol.~1,
  2018.

\bibitem{park2021learning}
H.~Park, S.~Lee, J.~Lee, B.~Ham, Learning by aligning: Visible-infrared person
  re-identification using cross-modal correspondences, in: International
  Conference on Computer Vision, 2021, pp. 12046--12055.

\bibitem{kang2019person}
J.~K. Kang, T.~M. Hoang, K.~R. Park, Person re-identification between visible
  and thermal camera images based on deep residual cnn using single input, IEEE
  Access 7 (2019) 57972--57984.

\bibitem{kniaz2018thermalgan}
V.~V. Kniaz, V.~A. Knyaz, J.~Hladuvka, W.~G. Kropatsch, V.~Mizginov,
  Thermalgan: Multimodal color-to-thermal image translation for person
  re-identification in multispectral dataset, in: European Conference on
  Computer Vision (ECCV) Workshops, 2018, pp. 0--0.

\bibitem{yolov5}
G.~Jocher, {Yolov5}, \url{https://github.com/ultralytics/yolov5} (Oct. 2020).

\bibitem{romberg2011scalable}
S.~Romberg, L.~G. Pueyo, R.~Lienhart, R.~Van~Zwol, Scalable logo recognition in
  real-world images, in: ACM International Conference on Multimedia Retrieval,
  2011, pp. 1--8.

\bibitem{he2016deep}
K.~He, X.~Zhang, S.~Ren, J.~Sun, Deep residual learning for image recognition,
  in: Conference on computer vision and pattern recognition, 2016.

\bibitem{niu2016ordinal}
Z.~Niu, M.~Zhou, L.~Wang, X.~Gao, G.~Hua, Ordinal regression with multiple
  output cnn for age estimation, in: Computer vision and pattern recognition,
  2016, pp. 4920--4928.

\bibitem{chavdarova2018wildtrack}
T.~Chavdarova, P.~Baqu{\'e}, S.~Bouquet, A.~Maksai, C.~Jose, T.~Bagautdinov,
  L.~Lettry, P.~Fua, L.~Van~Gool, F.~Fleuret, Wildtrack: A multi-camera hd
  dataset for dense unscripted pedestrian detection, in: Conference on Computer
  Vision and Pattern Recognition, 2018, pp. 5030--5039.

\bibitem{dendorfer2021motchallenge}
P.~Dendorfer, A.~Osep, A.~Milan, K.~Schindler, D.~Cremers, I.~Reid, S.~Roth,
  L.~Leal-Taix{\'e}, Motchallenge: A benchmark for single-camera multiple
  target tracking, International Journal of Computer Vision 129~(4).

\bibitem{ge2019deepfashion2}
Y.~Ge, R.~Zhang, X.~Wang, X.~Tang, P.~Luo, Deepfashion2: A versatile benchmark
  for detection, pose estimation, segmentation and re-identification of
  clothing images, in: Conference on Computer Vision and Pattern Recognition,
  2019, pp. 5337--5345.

\bibitem{hrkac2016tattoo}
T.~Hrkac, K.~Brkic, Z.~Kalafatic, Tattoo detection for soft biometric
  de-identification based on convolutional neural networks, in: Proc. OAGM-ARW
  Joint Workshop, 2016, pp. 131--138.

\bibitem{aytar2016soundnet}
Y.~Aytar, C.~Vondrick, A.~Torralba, Soundnet: Learning sound representations
  from unlabeled video, Advances in neural information processing systems 29.

\bibitem{krizhevsky2012imagenet}
A.~Krizhevsky, I.~Sutskever, G.~E. Hinton, Imagenet classification with deep
  convolutional neural networks, Advances in neural information processing
  systems 25.

\bibitem{zhou2014learning}
B.~Zhou, A.~Lapedriza, J.~Xiao, A.~Torralba, A.~Oliva, Learning deep features
  for scene recognition using places database, Advances in neural information
  processing systems 27.

\bibitem{wojke2017simple}
N.~Wojke, A.~Bewley, D.~Paulus, Simple online and realtime tracking with a deep
  association metric, in: International conference on image processing (ICIP),
  IEEE, 2017, pp. 3645--3649.

\bibitem{hochreiter1997long}
S.~Hochreiter, J.~Schmidhuber, Long short-term memory, Neural computation 9~(8)
  (1997) 1735--1780.

\bibitem{fernando2018soft+}
T.~Fernando, S.~Denman, S.~Sridharan, C.~Fookes, Soft+ hardwired attention: An
  lstm framework for human trajectory prediction and abnormal event detection,
  Neural networks 108 (2018) 466--478.

\bibitem{fernando2018pedestrian}
T.~Fernando, S.~Denman, S.~Sridharan, C.~Fookes, Pedestrian trajectory
  prediction with structured memory hierarchies, in: Joint European Conference
  on Machine Learning and Knowledge Discovery in Databases, Springer, 2018, pp.
  241--256.

\bibitem{kingma2014adam}
D.~P. Kingma, J.~Ba, Adam: A method for stochastic optimization, arXiv preprint
  arXiv:1412.6980.

\bibitem{sandler2018mobilenetv2}
M.~Sandler, A.~Howard, M.~Zhu, A.~Zhmoginov, L.-C. Chen, Mobilenetv2: Inverted
  residuals and linear bottlenecks, in: Conference on computer vision and
  pattern recognition, 2018, pp. 4510--4520.

\bibitem{hirzer2011person}
M.~Hirzer, C.~Beleznai, P.~M. Roth, H.~Bischof, Person re-identification by
  descriptive and discriminative classification, in: Scandinavian conference on
  Image analysis, Springer, 2011, pp. 91--102.

\bibitem{zheng2016mars}
L.~Zheng, Z.~Bie, Y.~Sun, J.~Wang, C.~Su, S.~Wang, Q.~Tian, Mars: A video
  benchmark for large-scale person re-identification, in: European conference
  on computer vision, Springer, 2016, pp. 868--884.

\bibitem{felzenszwalb2010object}
P.~F. Felzenszwalb, R.~B. Girshick, D.~McAllester, D.~Ramanan, Object detection
  with discriminatively trained part-based models, IEEE transactions on pattern
  analysis and machine intelligence 32~(9) (2010) 1627--1645.

\bibitem{dehghan2015gmmcp}
A.~Dehghan, S.~Modiri~Assari, M.~Shah, Gmmcp tracker: Globally optimal
  generalized maximum multi clique problem for multiple object tracking, in:
  Computer vision and pattern recognition, 2015, pp. 4091--4099.

\bibitem{chowdhury2018msu}
A.~Chowdhury, Y.~Atoum, L.~Tran, X.~Liu, A.~Ross, Msu-avis dataset: Fusing face
  and voice modalities for biometric recognition in indoor surveillance videos,
  in: International Conference on Pattern Recognition (ICPR), IEEE, 2018, pp.
  3567--3573.

\bibitem{sundararajan2017axiomatic}
M.~Sundararajan, A.~Taly, Q.~Yan, Axiomatic attribution for deep networks, in:
  International conference on machine learning, PMLR, 2017, pp. 3319--3328.

\bibitem{fernando2018soft}
T.~Fernando, S.~Denman, S.~Sridharan, C.~Fookes, Soft+ hardwired attention: An
  lstm framework for human trajectory prediction and abnormal event detection,
  Neural networks 108 (2018) 466--478.

\bibitem{priyasad2022affect}
D.~Priyasad, T.~Fernando, S.~Denman, S.~Sridharan, C.~Fookes, Affect
  recognition from scalp-eeg using channel-wise encoder networks coupled with
  geometric deep learning and multi-channel feature fusion, Knowledge-Based
  Systems (2022) 109038.

\bibitem{liao2013graph}
W.~Liao, R.~Bellens, A.~Pizurica, S.~Gautama, W.~Philips, Graph-based feature
  fusion of hyperspectral and lidar remote sensing data using morphological
  features, in: Geoscience and Remote Sensing Symposium, 2013.

\bibitem{priyasad2021memory}
D.~Priyasad, T.~Fernando, S.~Denman, S.~Sridharan, C.~Fookes, Memory based
  fusion for multi-modal deep learning, Information Fusion 67.

\bibitem{gammulle2019forecasting}
H.~Gammulle, S.~Denman, S.~Sridharan, C.~Fookes, Forecasting future action
  sequences with neural memory networks, British Machine Vision Conference.

\end{thebibliography}

\end{document}